\documentclass[10pt,onecolumn]{article}
 \usepackage{times}
\usepackage[dvips]{graphicx}
\usepackage[hang,scriptsize]{subfigure}
\usepackage{latexsym}

\usepackage{amsmath} 
\usepackage{amssymb} 

\makeatletter
\def\@maketitle{%
  \vbox to 6cm{
    \hsize\textwidth
    \linewidth\hsize
    \vspace{1.5cm}
    \centering
    {\bfseries\LARGE \@title \par}
    \vspace{12pt}
    {\fontsize{11pt}{13pt}\selectfont \begin{tabular}[t]{c}\@author \end{tabular}\par}
    \vfill} 
}
\renewcommand\section{\@startsection{section}{1}{\z@}%
                       {-12\p@ \@plus -4\p@ \@minus -4\p@}%
                       {6\p@ \@plus 4\p@ \@minus 4\p@}%
                       {\normalfont\large\bfseries
                        \rightskip=\z@ \@plus 8em\pretolerance=10000 }}
\renewcommand\subsection{\@startsection{subsection}{2}{\z@}%
                       {-12\p@ \@plus -4\p@ \@minus -4\p@}%
                       {6\p@ \@plus 4\p@ \@minus 4\p@}%
                       {\normalfont\fontsize{11pt}{13pt}\selectfont\bfseries
                        \rightskip=\z@ \@plus 8em\pretolerance=10000 }}
\renewcommand\subsubsection{\@startsection{subsubsection}{3}{\z@}%
                       {-12\p@ \@plus -4\p@ \@minus -4\p@}%
                       {6\p@ \@plus 4\p@ \@minus 4\p@}%
                       {\normalfont\normalsize\itshape}}
\renewcommand\paragraph{\@startsection{paragraph}{4}{\z@}%
                       {-12\p@ \@plus -4\p@ \@minus -4\p@}%
                       {-0.5em \@plus -0.22em \@minus -0.1em}%
                       {\normalfont\normalsize\itshape}}
\makeatother

\renewenvironment{abstract}%
  {\small
    \list{}{\labelwidth0pt
      \leftmargin0pt \rightmargin\leftmargin
      \listparindent\parindent \itemindent0pt
      \parsep0pt
      }%
    \item[\hskip\labelsep\bfseries\abstractname\enspace --] \itshape}{\endlist}

\newcommand{\keywordsname}{Keywords}
\newenvironment{keywords}%
  {\small
    \list{}{\labelwidth0pt
      \leftmargin0pt \rightmargin\leftmargin
      \listparindent\parindent \itemindent0pt
      \parsep0pt
      }%
    \item[\hskip\labelsep\bfseries\keywordsname:]}{\endlist}

\begin{document}

\title{An Introduction to the DSm Theory for the Combination\\ 
of Paradoxical, Uncertain, and Imprecise Sources of Information\thanks{Online paper revised on August 1st, 2006.}}

\author{\begin{tabular}{c@{\extracolsep{4em}}c@{\extracolsep{1em}}c}
{\bf Florentin Smarandache} & {\bf Jean Dezert}\\
Dept. of Mathematics &  ONERA/DTIM/IED \\
Univ. of New Mexico  & 29 Av. de la  Division Leclerc \\
Gallup, NM 8730 & 92320 Ch\^{a}tillon \\
U.S.A.  & France \\
{\tt smarand@unm.edu} & {\tt Jean.Dezert@onera.fr}
\end{tabular}}
\vspace{-2cm}

\date{Aug 1, 2006}
\maketitle
\thispagestyle{empty}
\pagestyle{empty}

\begin{abstract}
The management and combination of uncertain, imprecise, fuzzy and even paradoxical or high conflicting sources of information has always been, and still remains today, of primal importance for the development of reliable modern information systems involving artificial reasoning. In this introduction, we present a survey of our recent theory of plausible and paradoxical reasoning, known as Dezert-Smarandache Theory (DSmT) in the literature, developed for dealing with imprecise, uncertain and paradoxical sources of information. We focus our presentation here rather on the foundations of DSmT, and on the two important new rules of combination, than on browsing specific applications of DSmT available in literature. Several simple examples are given throughout the presentation to show the efficiency and the generality of this new approach.
\end{abstract}


\section{Introduction}

The management and combination of uncertain, imprecise, fuzzy and even paradoxical or high conflicting sources of information has always been, and still remains today, of primal importance for the development of reliable modern information systems involving artificial reasoning. The combination (fusion) of information arises in many fields of applications nowadays (especially in defense, medicine, finance, geo-science, economy, etc). When several sensors, observers or experts have to be combined together to solve a problem, or if one wants to update our current estimation of solutions for a given problem with some new information available, we need powerful and solid mathematical tools for the fusion, specially when the information one has to deal with is imprecise and uncertain. In this paper, we present a survey of our recent theory of plausible and paradoxical reasoning, known as Dezert-Smarandache Theory (DSmT) in the literature, developed for dealing with imprecise, uncertain and paradoxical sources of information. Recent publications have shown the interest and the ability of DSmT to solve problems where other approaches fail, especially when conflict between sources becomes high. We focus our presentation here rather on the foundations of DSmT, and on the two important new rules of combination, than on browsing specific applications of DSmT available in literature. A particular attention is given to general (hybrid) rule of combination which deals with any model for fusion problems, depending on the nature of elements or hypotheses involved into them. The Shafer's model on which is based the Dempster-Shafer Theory (DST) appears only as a specific DSm hybrid model and can be easily handled by our approach as well. Several simple examples are given throughout the presentation to show the efficiency and the generality of this new approach.

\section{Foundations of the  DSmT}

The development of the DSmT (Dezert-Smarandache Theory of plausible and paradoxical reasoning \cite{DSmTBook_2004a,Dezert_2006a}) arises from the necessity to overcome the inherent limitations of the DST (Dempster-Shafer Theory \cite{Shafer_1976}) which are closely related with the acceptance of Shafer's model for the fusion problem under consideration (i.e. the frame of {\it{discernment}} $\Theta$ defined as a finite set of {\it{exhaustive}} and {\it{exclusive}} hypotheses $\theta_i$, $i=1,\ldots,n$), the third middle excluded principle (i.e. the existence of the complement for any elements/propositions belonging to the power set of $\Theta$), and the acceptance of Dempter's rule of combination (involving normalization) as the framework for the combination of independent sources of evidence. Discussions on limitations of DST and presentation of some alternative rules to the Dempster's rule of combination can be found in \cite{Zadeh_1979,Zadeh_1984,Zadeh_1985,Yager_1985,Zadeh_1986,Dubois_1986c,Yager_1987,Pearl_1988,Smets_1988,Voorbraak_1991,Inagaki_1991,Murphy_2000,Lefevre_2002,Sentz_2002,Lefevre_2003,DSmTBook_2004a} and therefore they will be not reported in details in this paper. We argue that these three fundamental conditions of the DST can be removed and another new mathematical approach for combination of evidence is possible.\\

The basis of the DSmT is  the refutation of the principle of the third excluded middle and Shafer's model, since for a wide class of fusion problems the intrinsic nature of hypotheses can be only vague and imprecise in such a way that precise refinement is just impossible to obtain in reality so that the exclusive elements $\theta_i$ cannot be properly identified and precisely separated. Many problems involving fuzzy continuous and relative concepts described in natural language and having no absolute interpretation like tallness/smallness, pleasure/pain, cold/hot, Sorites paradoxes, etc,  enter in this category. DSmT starts with the notion of {\it{free DSm model}}, denoted $\mathcal{M}^f(\Theta)$, and considers $\Theta$ only as a frame of exhaustive elements $\theta_i$, $i=1,\ldots,n$ which can potentially overlap. This model is {\it{free}} because no other assumption is done on the hypotheses, but the weak exhaustivity constraint which can always been satisfied according the closure principle explained in \cite{DSmTBook_2004a}. No other constraint is involved in the free DSm model.
When the free DSm model holds, the classic commutative and associative DSm rule of combination (corresponding to the conjunctive consensus defined on the free Dedekind's lattice) is performed.\\

Depending on the intrinsic nature of the elements of the fusion problem under consideration, it can however happen that the free model does not fit the reality because some subsets of $\Theta$ can contain elements known to be truly exclusive but also truly non existing at all at a given time (specially when working on dynamic fusion problem where the frame $\Theta$ varies with time with the revision of the knowledge available). These integrity constraints are then explicitly and formally introduced into the free DSm model $\mathcal{M}^f(\Theta)$ in order to adapt it properly to fit as close as possible with the reality and permit to construct a {\it{hybrid DSm model}} $\mathcal{M}(\Theta)$ on which the combination will be efficiently performed. Shafer's model, denoted $\mathcal{M}^0(\Theta)$, corresponds to a very specific hybrid DSm model including all possible exclusivity constraints. The DST has been developed for working only with $\mathcal{M}^0(\Theta)$ while the DSmT has been developed for working with any kind of hybrid model (including Shafer's model and the free DSm model), to manage as efficiently and precisely as possible imprecise, uncertain and potentially high conflicting sources of evidence while keeping in mind the possible dynamicity of the information fusion problematic. The foundations of the DSmT are therefore totally different from those of all existing approaches managing uncertainties, imprecisions and conflicts. DSmT provides a new interesting way to attack the  information fusion problematic with a general framework in order to cover a wide variety of problems. \\

DSmT refutes also the idea that sources of evidence provide their beliefs with the same absolute interpretation of elements of the same frame $\Theta$ and the conflict between sources arises not only because of the possible unreliabilty of sources, but also because of possible different and relative interpretation of $\Theta$, e.g. what is considered as good for somebody can be considered as bad for somebody else. There is some unavoidable subjectivity in the belief assignments provided by the sources of evidence, otherwise it would mean that all bodies of evidence have a same objective and universal interpretation (or measure) of the phenomena under consideration, which unfortunately rarely occurs in reality, but when bba are based on some {\it{objective probabilities}} transformations. But in this last case, probability theory can handle properly and efficiently the information, and the DST, as well as the DSmT, becomes useless. If we now get out of the probabilistic background argumentation for the construction of bba, we claim that in most of cases, the sources of evidence provide their beliefs about elements of the frame of the fusion problem only based on their own limited knowledge and experience without reference to the (inaccessible) absolute truth of the space of possibilities. First applications of DSmT for target tracking, satellite surveillance, situation analysis and sensor allocation optimization can be found in \cite{DSmTBook_2004a}.

\subsection{Notion of hyper-power set $D^\Theta$}

One of the cornerstones of the DSmT is  the free Dedekind lattice \cite{Dedekind_1897} denoted {\it{hyper-power set}} in the DSmT framework. Let $\Theta=\{\theta_{1},\ldots,\theta_{n}\}$ be a finite set (called frame) of $n$ 
exhaustive elements\footnote{We do not assume here that elements $\theta_i$ are necessary exclusive. There is no restriction on $\theta_i$ but the exhaustivity.}. The hyper-power set $D^\Theta$ is defined as the set of all composite propositions built from elements of $\Theta$ with $\cup$ and $\cap$ operators\footnote{$\Theta$ generates $D^\Theta$ under operators $\cup$ and $\cap$} such that: 
\begin{enumerate}
\item $\emptyset, \theta_1,\ldots, \theta_n \in D^\Theta$.
\item  If $A, B \in D^\Theta$, then $A\cap B\in D^\Theta$ and $A\cup B\in D^\Theta$.
\item No other elements belong to $D^\Theta$, except those obtained by using rules 1 or 2.
\end{enumerate}
The dual (obtained by switching $\cup$ and $\cap$ in expressions) of $D^\Theta$ is itself.  There are elements in $D^\Theta$ which are self-dual (dual to themselves), for example $\alpha_8$ for the case when $n=3$ in the following example. The cardinality of $D^\Theta$ is majored by 
$2^{2^n}$ when the cardinality of $\Theta$ equals $n$, i.e. $\vert\Theta\vert=n$. The generation 
of hyper-power set $D^\Theta$ is closely related with the famous Dedekind's problem \cite{Dedekind_1897,Comtet_1974} on enumerating the 
set of isotone Boolean functions. The generation of the hyper-power set is presented in \cite{DSmTBook_2004a}. Since for any given finite set $\Theta$, $\vert D^\Theta\vert  \geq \vert 2^\Theta\vert $ we call $D^\Theta$ the  {\it{hyper-power set}} of $\Theta$.\\

\noindent{\it{Example of the first hyper-power sets $D^\Theta$}}
\begin{itemize}
\item
For the degenerate case ($n=0)$ where $\Theta=\{  \}$, one has $D^\Theta=\{\alpha_0\triangleq\emptyset\}$ and $\vert D^\Theta\vert = 1$.
\item When $\Theta=\{\theta_{1}\}$, one has $D^\Theta=\{\alpha_0\triangleq\emptyset,\alpha_1\triangleq\theta_1 \}$ and $\vert D^\Theta\vert = 2$.
\item When $\Theta=\{\theta_{1},\theta_{2}\}$, one has
$D^\Theta=\{\alpha_0,\alpha_1,\ldots,\alpha_{4} \}$ and $\vert D^\Theta\vert = 5$ with
$\alpha_0\triangleq\emptyset$, $\alpha_1\triangleq\theta_1\cap\theta_2$, $\alpha_2\triangleq\theta_1$, $\alpha_3\triangleq\theta_2 $ and $\alpha_4\triangleq\theta_1\cup\theta_2 $.
\item
When $\Theta=\{\theta_{1},\theta_{2},\theta_{3}\}$,  one has $D^\Theta=\{\alpha_0,\alpha_1,\ldots,\alpha_{18} \}$ and $\vert D^\Theta\vert = 19$ with
\begin{equation*}
\begin{array}{ll}
\alpha_0\triangleq\emptyset          &                                       \\
\alpha_1\triangleq\theta_1\cap\theta_2\cap\theta_3    &\alpha_{10}\triangleq\theta_2   \\
\alpha_2\triangleq\theta_1\cap\theta_2    & \alpha_{11}\triangleq\theta_3                            \\
\alpha_3\triangleq\theta_1\cap\theta_3    &\alpha_{12}\triangleq(\theta_1\cap\theta_2)\cup\theta_3                            \\
\alpha_4\triangleq\theta_2\cap\theta_3    &\alpha_{13}\triangleq(\theta_1\cap\theta_3)\cup\theta_2                             \\
\alpha_5\triangleq(\theta_1\cup\theta_2)\cap\theta_3  & \alpha_{14}\triangleq(\theta_2\cap\theta_3)\cup\theta_1      \\
\alpha_6\triangleq(\theta_1\cup\theta_3)\cap\theta_2  & \alpha_{15}\triangleq\theta_1\cup\theta_2       \\
\alpha_7\triangleq(\theta_2\cup\theta_3)\cap\theta_1 & \alpha_{16}\triangleq\theta_1\cup\theta_3        \\
\alpha_8\triangleq(\theta_1\cap\theta_2)\cup(\theta_1\cap\theta_3)\cup(\theta_2\cap\theta_3)  & \alpha_{17}\triangleq\theta_2\cup\theta_3    \\
\alpha_9\triangleq\theta_1  & \alpha_{18}\triangleq\theta_1\cup\theta_2\cup\theta_3                                                   \end{array}
\end{equation*}
\end{itemize}

The cardinality of hyper-power set $D^\Theta$ for $n\geq1$ follows the sequence of Dedekind's numbers \cite{Sloane_2003}, i.e.
1,2,5,19,167, 7580,7828353,...  and analytical expression of Dedekind's numbers has been obtained recently by Tombak in \cite{Tombak_2001} (see \cite{DSmTBook_2004a} for details on generation and ordering of $D^\Theta$).

\subsection{Notion of free and hybrid DSm models}
\label{Sec:DSMmodels}

Elements $\theta_i$, $i=1,\ldots,n$ of $\Theta$ constitute the finite set of hypotheses/concepts characterizing the fusion problem under consideration. $D^\Theta$ constitutes what we call the {\it{free DSm model}} $\mathcal{M}^f(\Theta)$ and allows to work with fuzzy concepts which depict a continuous and relative intrinsic nature. Such kinds of concepts cannot be precisely refined in an absolute interpretation because of the unapproachable universal truth.\\

However for some particular fusion problems involving discrete concepts, elements $\theta_i$ are truly exclusive. In such case, all the exclusivity constraints on $\theta_i$, $i=1,\ldots,n$ have to be included in the previous model to characterize properly the true nature of the fusion problem and to fit it with the reality. By doing this, the hyper-power set $D^\Theta$ reduces naturally to the classical power set $2^\Theta$ and this constitutes the most restricted hybrid DSm model, denoted  $\mathcal{M}^0(\Theta)$, coinciding with Shafer's model. As an exemple, let's consider the 2D problem where $\Theta=\{\theta_1,\theta_2\}$ with $D^\Theta=\{\emptyset,\theta_1\cap\theta_2,\theta_1,\theta_2,\theta_1\cup\theta_2\}$ and assume now that $\theta_1$ and $\theta_2$ are truly exclusive (i.e. Shafer's model $\mathcal{M}^0$ holds), then because $\theta_1\cap\theta_2\overset{\mathcal{M}^0}{=}\emptyset$, one gets $D^\Theta=\{\emptyset,\theta_1\cap\theta_2\overset{\mathcal{M}^0}{=}\emptyset,\theta_1,\theta_2,\theta_1\cup\theta_2\}=\{\emptyset,\theta_1,\theta_2,\theta_1\cup\theta_2\}\equiv 2^\Theta$.\\

Between the class of fusion problems corresponding to the free DSm model $\mathcal{M}^f(\Theta)$ and  the class of fusion problems corresponding to Shafer's model $\mathcal{M}^0(\Theta)$, there exists another wide class of hybrid fusion problems involving in $\Theta$ both fuzzy continuous concepts and discrete hypotheses. In such (hybrid) class, some exclusivity constraints and possibly some non-existential constraints (especially when working on dynamic\footnote{i.e. when the frame $\Theta$ and/or the model $\mathcal{M}$ is changing with time.} fusion) have to be taken into account. Each hybrid fusion problem of this class will then be characterized by a proper hybrid DSm model $\mathcal{M}(\Theta)$ with $\mathcal{M}(\Theta)\neq\mathcal{M}^f(\Theta)$ and $\mathcal{M}(\Theta)\neq \mathcal{M}^0(\Theta)$. As simple example of DSm hybrid model, let's consider the 3D case with the frame $\Theta=\{\theta_1,\theta_2,\theta_3\}$ with the model $\mathcal{M}\neq\mathcal{M}^f$ in which we force all possible conjunctions to be empty, but $\theta_1\cap\theta_2$. This hybrid DSm model is then represented with  the following Venn diagram (where boundaries of intersection of $\theta_1$ and $\theta_2$ are not precisely defined if $\theta_1$ and $\theta_2$ represent only fuzzy concepts like {\it{smallness}} and {\it{tallness}} by example).
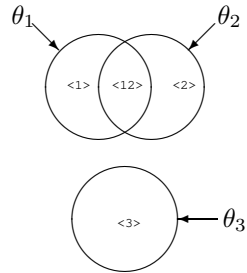
\begin{center}
{\tt \setlength{\unitlength}{1pt}
\begin{picture}(90,90)
\thinlines    
\put(40,60){\circle{40}}
\put(60,60){\circle{40}}
\put(50,10){\circle{40}}
\put(15,84){\vector(1,-1){10}}
\put(7,84){$\theta_{1}$}
\put(84,84){\vector(-1,-1){10}}
\put(85,84){$\theta_{2}$}
\put(85,10){\vector(-1,0){15}}
\put(87,7){$\theta_{3}$}
\put(45,59){\tiny{<12>}}
\put(47,7){\tiny{<3>}}
\put(68,59){\tiny{<2>}}
\put(28,59){\tiny{<1>}}
\end{picture}}
\end{center}

\vspace{1cm}

\subsection{Generalized belief functions}

From a general frame $\Theta$, we define a map $m(.): 
D^\Theta \rightarrow [0,1]$ associated to a given body of evidence $\mathcal{B}$ as 
\begin{equation}
m(\emptyset)=0 \qquad \text{and}\qquad \sum_{A\in D^\Theta} m(A) = 1 
\end{equation}
\noindent The quantity $m(A)$ is called the {\it{generalized basic belief assignment/mass}} (gbba) of $A$.\\

\noindent
The {\it{generalized belief and plausibility functions}} are defined in almost the same manner as within the DST, i.e.
\begin{equation}
\text{Bel}(A) = \sum_{\substack{B\subseteq A\\ B\in D^\Theta}} m(B)
\qquad\qquad\text{Pl}(A) = \sum_{\substack{B\cap A\neq\emptyset \\ B\in D^\Theta}} m(B)
\end{equation}

These definitions are compatible with the definitions of classical belief functions in the DST framework when $D^\Theta$ reduces to $2^\Theta$ for fusion problems where Shafer's model $\mathcal{M}^0(\Theta)$ holds. We still have $\forall A\in D^\Theta,\, \text{Bel}(A)\leq \text{Pl}(A)$.
Note that when working with the free DSm model $\mathcal{M}^f(\Theta)$, one has always $\text{Pl}(A) =1$ $\forall A\neq\emptyset \in D^\Theta$ which is normal.

\subsection{The classic DSm rule of combination}

When the free DSm model $\mathcal{M}^f(\Theta)$ holds for the fusion problem under consideration, the  classic DSm rule of combination $m_{\mathcal{M}^f(\Theta)}\equiv m(.)\triangleq [m_{1}\oplus m_{2}](.)$ of two independent\footnote{While independence is a difficult concept to define in all theories managing epistemic uncertainty, we follow here the interpretation of Smets in \cite{Smets_1986} and \cite{Smets_1988}, p. 285 and consider that two sources of evidence are independent (i.e distinct and noninteracting) if each leaves one totally ignorant about the particular value the other will take.} sources of evidences $\mathcal{B}_{1}$ and  $\mathcal{B}_{2}$ over the same 
frame $\Theta$ with belief functions $\text{Bel}_{1}(.)$ and 
 $\text{Bel}_{2}(.)$ associated with gbba $m_{1}(.)$ and $m_{2}(.)$ corresponds to the conjunctive consensus of the sources. It is  given by \cite{DSmTBook_2004a}:
 \begin{equation}
\forall C\in D^\Theta,\qquad m_{\mathcal{M}^f(\Theta)}(C) \equiv m(C) = 
 \sum_{\substack{A,B\in D^\Theta\\ A\cap B=C}}m_{1}(A)m_{2}(B)
 \label{JDZT}
 \end{equation}
 
Since $D^\Theta$ is closed under $\cup$ and $\cap$ set operators, this new rule 
of combination guarantees that $m(.)$ is a proper generalized belief assignment, i.e. $m(.): D^\Theta \rightarrow [0,1]$. This rule of combination is commutative and associative 
and can always be used for the fusion of sources involving fuzzy concepts when free DSm model holds for the problem under consideration. This rule can be directly and easily extended for the combination of $k > 2$ independent sources of evidence \cite{DSmTBook_2004a}.\\

This classic DSm rule of combination looks very expensive in terms of computations and memory size due to the huge number of elements in $D^\Theta$ when the cardinality of $\Theta$ increases. This remark is however valid only if the cores (the set of focal elements of gbba) $\mathcal{K}_1(m_1)$ and $\mathcal{K}_2(m_2)$ coincide with $D^\Theta$, i.e. when $m_1(A)>0$ and $m_2(A)>0$ for all $A\neq\emptyset\in D^\Theta$. Fortunately, it is important to note here that in most of the practical applications the sizes of $\mathcal{K}_1(m_1)$ and $\mathcal{K}_2(m_2)$ are much smaller than $\vert D^\Theta\vert$ because bodies of evidence generally allocate their basic belief assignments only over a subset of the hyper-power set. This makes things easier for the implementation of the classic DSm rule \eqref{JDZT}. The DSm rule is actually very easy to implement. It suffices for each focal element of $\mathcal{K}_1(m_1)$ to multiply it with the focal elements of $\mathcal{K}_2(m_2)$ and then to  pool all combinations which are equivalent under the  algebra of sets.\\

While very costly in term on merory storage in the worst case (i.e. when all $m(A)>0$, $A\in D^\Theta$ or $A\in2^{\Theta_{ref}}$), the DSm rule however requires much smaller memory storage than for the DST working on the ultimate refinement $2^{\Theta_{ref}}$ of same initial frame $\Theta$ as shown in following table
\begin{center}
\begin{tabular}{|l|l|l|}
\hline
$\vert\Theta\vert=n$ & $\vert D^\Theta \vert$ & $\vert 2^{\Theta_{ref}} \vert = 2^{2^n-1}$ \\
\hline
2 &  5 &  $2^3=8$ \\
3 &  19 &  $2^7=128$ \\
4 &  167 &  $2^{15}=32768$ \\
5 &  7580 &  $2^{31}=2147483648$\\
\hline
\end{tabular}
\end{center}
 
However in most fusion applications only a small subset of elements of $D^\Theta$ have a non null basic belief mass because all the commitments are just usually impossible to assess precisely when the dimension of the problem increases. Thus, it is not necessary to generate and keep in memory all elements of $D^\Theta$ or $2^{\Theta_{ref}}$ but only those which have a positive belief mass.  However there is a real technical challenge on how to manage efficiently all elements of the hyper-power set. This problem is obviously much more difficult when trying to work on the refined frame of discernment $2^{\Theta_{ref}}$ if one prefer to apply Dempster-Shafer theory and use the Dempster's rule of combination. It is important to keep in mind that the ultimate refined frame consisting in exhaustive and exclusive finite set of refined hypotheses is just impossible to justify and to define precisely for all problems dealing with fuzzy and ill-defined continuous concepts. A full discussion and example on refinement can be found in \cite{DSmTBook_2004a}.

\subsection{The hybrid DSm rule of combination}

When the free DSm model $\mathcal{M}^f(\Theta)$ does not hold due to the true nature of the fusion problem under consideration which requires to take into account some known integrity constraints, one has to work with a proper hybrid DSm model $\mathcal{M}(\Theta)\neq\mathcal{M}^f(\Theta)$. In such case, the hybrid DSm rule of combination based on the chosen hybrid DSm model $\mathcal{M}(\Theta)$ for $k\geq 2$ independent sources of information is defined for all $A\in D^\Theta$ as \cite{DSmTBook_2004a}:
\begin{equation}
m_{\mathcal{M}(\Theta)}(A)\triangleq 
\phi(A)\Bigl[ S_1(A) + S_2(A) + S_3(A)\Bigr]
 \label{eq:DSmHkBis1}
\end{equation}
\noindent
where all sets involved in formulas are in the canonical form and $\phi(A)$ is the {\it{characteristic non-emptiness function}} of a set $A$, i.e. $\phi(A)= 1$ if  $A\notin \boldsymbol{\emptyset}$ and $\phi(A)= 0$ otherwise, where $\boldsymbol{\emptyset}\triangleq\{\boldsymbol{\emptyset}_{\mathcal{M}},\emptyset\}$. $\boldsymbol{\emptyset}_{\mathcal{M}}$ is the set  of all elements of $D^\Theta$ which have been forced to be empty through the constraints of the model $\mathcal{M}$ and $\emptyset$ is the classical/universal empty set. $S_1(A)\equiv m_{\mathcal{M}^f(\theta)}(A)$, $S_2(A)$, $S_3(A)$ are defined by 
\begin{equation}
S_1(A)\triangleq \sum_{\substack{X_1,X_2,\ldots,X_k\in D^\Theta\\ X_1\cap X_2\cap\ldots\cap X_k=A}} \prod_{i=1}^{k} m_i(X_i)
\end{equation}
\begin{equation}
S_2(A)\triangleq \sum_{\substack{X_1,X_2,\ldots,X_k\in\boldsymbol{\emptyset}\\ [\mathcal{U}=A]\vee [(\mathcal{U}\in\boldsymbol{\emptyset}) \wedge (A=I_t)]}} \prod_{i=1}^{k} m_i(X_i)
\end{equation}
\begin{equation}
S_3(A)\triangleq\sum_{\substack{X_1,X_2,\ldots,X_k\in D^\Theta \\ X_1\cup X_2\cup\ldots\cup X_k=A \\ X_1\cap X_2\cap \ldots\cap X_k \in\boldsymbol{\emptyset}}}  \prod_{i=1}^{k} m_i(X_i)
\end{equation}
with $\mathcal{U}\triangleq u(X_1)\cup u(X_2)\cup \ldots \cup u(X_k)$ where $u(X)$ is the union of all $\theta_i$ that compose $X$, $I_t \triangleq \theta_1\cup \theta_2\cup\ldots\cup \theta_n$ is the total ignorance.
$S_1(A)$ corresponds to the classic DSm rule for $k$ independent sources based on the free DSm model $\mathcal{M}^f(\Theta)$; $S_2(A)$ represents the mass of all relatively and absolutely empty sets which is transferred to the total or relative ignorances associated with non existential constraints (if any, like in some dynamic problems); $S_3(A)$ transfers the sum of relatively empty sets directly onto the canonical disjunctive form of non-empty sets.\\

The hybrid DSm rule of combination generalizes the classic DSm rule of combination and is not equivalent to Dempter's rule. It works for any models (the free DSm model, Shafer's model or any other hybrid models) when manipulating {\it{precise}} generalized (or eventually classical) basic belief functions. An extension of this rule for the combination of {\it{imprecise}} generalized (or eventually classical) basic belief functions is presented in next section.\\

Note that in DSmT framework it is also possible to deal directly with complements if necessary depending on the problem under consideration and the information provided by the sources of evidence themselves. The first and simplest way is to work on Shafer's model when utimate refinement is possible.
The second way is to deal with partially known frame and introduce directly the complementary hypotheses into the frame itself. By example, if one knows only two hypotheses $\theta_1$, $\theta_2$ and their complements $\bar{\theta}_1$, $\bar{\theta}_2$, then can choose $\Theta=\{\theta_1,\theta_2,\bar{\theta}_1,\bar{\theta}_2\}$. In such case, we don't necessarily assume that $\bar{\theta}_1=\theta_2$ and $\bar{\theta}_2=\theta_1$ because $\bar{\theta}_1$ and $\bar{\theta}_2$ may include other unknown hypotheses we have no information about (case of partial known frame). More generally, in DSmT framework, it is not necessary that the frame is built on pure/simple (possibly vague) hypotheses $\theta_i$ as usually done in all theories managing uncertainty. The frame $\Theta$ can also contain directly as elements conjunctions and/or disjunctions (or mixed propositions) and negations/complements of pure hypotheses as well. The DSm rules also work in such non-classic frames because DSmT works on any distributive lattice built from $\Theta$ anywhere $\Theta$ is defined.

\subsection{Examples of combination rules}

Here are some numerical examples on results obtained by DSm rules of combination. More examples can be found in \cite{DSmTBook_2004a}.

\subsubsection{Example with $\Theta=\{\theta_1,\theta_2,\theta_3,\theta_4\}$}

Let's consider the frame of discernment $\Theta=\{\theta_1,\theta_2,\theta_3,\theta_4\}$, two independent experts, and the two following bbas
$$m_1(\theta_1)=0.6\quad m_1(\theta_3)=0.6\quad m_2(\theta_2)=0.6\quad m_2(\theta_4)=0.6$$
\noindent represented in terms of mass matrix 
\begin{equation*}
\mathbf{M}=
\begin{bmatrix}
0.6 & 0  & 0.4 & 0\\
0 & 0.2 & 0 & 0.8
\end{bmatrix}
\end{equation*}
\begin{itemize}
\item The Dempster's rule can not be applied because: $\forall 1\leq j \leq 4$, one gets $m(\theta_j) = 0/0$ (undefined!).
\item But the classic DSm rule works because one obtains: $m(\theta_1) = m(\theta_2) = m(\theta_3) = m(\theta_4) = 0$, 
and $m(\theta_1 \cap \theta_2) = 0.12$, $m(\theta_1 \cap \theta_4) = 0.48$,
$m(\theta_2 \cap \theta_3) = 0.08$, $m(\theta_3 \cap \theta_4) = 0.32$ (partial paradoxes/conflicts).
\item Suppose now one finds out that all intersections are empty (Shafer's model), then one applies the hybrid DSm rule and one gets (index $h$ stands here for {\it{hybrid}} rule): $m_h(\theta_1\cup \theta_2)=0.12$, $m_h(\theta_1\cup\theta_4)=0.48$, $m_h(\theta_2\cup\theta_3)=0.08$ and $m_h(\theta_3\cup\theta_4)=0.32$.
\end{itemize}

\subsubsection{Generalization of Zadeh's example with $\Theta=\{\theta_1,\theta_2,\theta_3\}$}

Let's consider $0 < \epsilon_1,\epsilon_2 < 1$ be two very tiny positive numbers (close to zero), the frame of discernment be $\Theta=\{\theta_1,\theta_2,\theta_3\}$, have two experts (independent sources of evidence $s_1$ and $s_2$) giving the belief masses
$$m_1(\theta_1)=1-\epsilon_1 \quad m_1(\theta_2)=0 \quad m_1(\theta_3)=\epsilon_1$$
$$m_2(\theta_1)=0 \quad m_2(\theta_2)=1-\epsilon_2 \quad m_2(\theta_3)=\epsilon_2$$
\noindent From now on, we prefer to use matrices to describe the masses, i.e.
$$\begin{bmatrix}
1-\epsilon_1 & 0 &\epsilon_1\\
0 & 1-\epsilon_2 & \epsilon_2
\end{bmatrix}
$$
\begin{itemize}
\item Using Dempster's rule of combination, one gets
$$m(\theta_3)=\frac{(\epsilon_1\epsilon_2)}{(1-\epsilon_1)\cdot 0 + 0\cdot (1-\epsilon_2) + \epsilon_1\epsilon_2}=1$$
\noindent which is absurd (or at least counter-intuitive). Note that whatever positive values for $\epsilon_1$, $\epsilon_2$ are, Dempster's rule of combination provides always the same result (one) which is abnormal. The only acceptable and correct result obtained by Dempster's rule is really obtained only in the trivial case when $\epsilon_1=\epsilon_2=1$, i.e. when both sources agree in $\theta_3$ with certainty which is obvious.
\item Using the DSm rule of combination based on free-DSm model, one gets 
$m(\theta_3)=\epsilon_1\epsilon_2$, $m(\theta_1\cap \theta_2)=(1-\epsilon_1)(1-\epsilon_2)$, $m(\theta_1\cap \theta_3)=(1-\epsilon_1)\epsilon_2$, $m(\theta_2\cap \theta_3)=(1-\epsilon_2)\epsilon_1$ and the others are zero which appears more reliable/trustable.
\item Going back to Shafer's model and using the hybrid DSm rule of combination, one gets 
$m(\theta_3)=\epsilon_1\epsilon_2$, $m(\theta_1\cup \theta_2)=(1-\epsilon_1)(1-\epsilon_2)$, $m(\theta_1\cup \theta_3)=(1-\epsilon_1)\epsilon_2$, $m(\theta_2\cup \theta_3)=(1-\epsilon_2)\epsilon_1$ and the others are zero.
\end{itemize}

\noindent
Note that in the special  case when $\epsilon_1=\epsilon_2=1/2$, one has
$$m_1(\theta_1)=1/2 \quad m_1(\theta_2)=0 \quad m_1(\theta_3)=1/2 \qquad\text{and}\qquad m_2(\theta_1)=0 \quad m_2(\theta_2)=1/2 \quad m_2(\theta_3)=1/2$$
Dempster's rule of combinations still yields $m(\theta_3)=1$ while the hybrid DSm rule based on the same Shafer's model yields now
$m(\theta_3)=1/4$, $m(\theta_1\cup \theta_2)=1/4$, $m(\theta_1\cup \theta_3)=1/4$, $m(\theta_2\cup \theta_3)=1/4$ which is normal.

\subsubsection{Comparison with Smets, Yager and Dubois \& Prade rules}

We compare the results provided by DSmT rules and the main common rules of combination on the following very simple numerical example where only 2 independent sources (a priori assumed equally reliable) are involved and providing their belief initially on the 3D frame $\Theta=\{\theta_1,\theta_2,\theta_3\}$. It is assumed in this example that Shafer's model holds and thus the belief assignments $m_1(.)$ and $m_2(.)$ do not  commit belief to internal conflicting information. $m_1(.)$ and $m_2(.)$ are chosen as follows:
 $$m_1(\theta_1)=0.1 \qquad m_1(\theta_2)=0.4 \qquad m_1(\theta_3)=0.2 \qquad m_1(\theta_1\cup \theta_2)=0.1$$
 $$m_2(\theta_1)=0.5 \qquad m_2(\theta_2)=0.1 \qquad m_2(\theta_3)=0.3 \qquad m_2(\theta_1\cup \theta_2)=0.1$$
 
\noindent
These belief masses are usually represented in the form of a belief mass matrix $\mathbf{M}$ given by
\begin{equation}
\mathbf{M}=
\begin{bmatrix}
0.1 & 0.4 & 0.2 & 0.3\\
0.5 & 0.1 & 0.3 & 0.1
\end{bmatrix}
\end{equation}
\noindent
where index $i$ for the rows corresponds to the index of the source no. $i$ and the indexes $j$ for columns of $\mathbf{M}$ correspond to a given choice for enumerating the focal elements of all sources.
In this particular example, index $j=1$ corresponds to $\theta_1$, $j=2$ corresponds to $\theta_2$, $j=3$ corresponds to $\theta_3$ and  $j=4$ corresponds to $\theta_1\cup \theta_2$.\\

Now let's imagine that one finds out that $\theta_3$ is actually truly empty because some extra and certain knowledge on $\theta_3$ is received by the fusion center. As example, $\theta_1$, $\theta_2$ and $\theta_3$ may correspond to three suspects (potential murders) in a police investigation, $m_1(.)$ and $m_2(.)$ corresponds to two reports of independent witnesses, but it turns out that finally $\theta_3$ has provided a strong alibi to the criminal police investigator once arrested by the policemen. This situation corresponds to set up a hybrid model $\mathcal{M}$ with the constraint $\theta_3\overset{\mathcal{M}}{=}\emptyset$. \\

Let's examine the result of the fusion in such situation obtained by the Smets', Yager's, Dubois \& Prade's and hybrid DSm rules of combinations. First note that, based on the free DSm model, one would get by applying the classic DSm rule (denoted here by index $DSmc$) the following fusion result

\begin{align*}
m_{DSmc}(\theta_1)&=0.21 \qquad m_{DSmc}(\theta_2)=0.11 \qquad m_{DSmc}(\theta_3)=0.06 \qquad m_{DSmc}(\theta_1\cup\theta_2)=0.03\\
m_{DSmc}(\theta_1\cap\theta_2)&=0.21 \qquad m_{DSmc}(\theta_1\cap\theta_3)=0.13\qquad m_{DSmc}(\theta_2\cap\theta_3)=0.14\\
m_{DSmc}(\theta_3\cap(\theta_1\cup\theta_2))&=0.11\\
\end{align*}

But because of the exclusivity constraints (imposed here by the use of Shafer's model and by the non-existential constraint $\theta_3\overset{\mathcal{M}}{=}\emptyset$), the total conflicting mass is actually given by
$$k_{12}=0.06 + 0.21 + 0.13 + 0.14 + 0.11=0.65 \qquad\text{(conflicting mass)}$$
\begin{itemize}
\item If one applies  {\bf{Dempster's rule}} \cite{Shafer_1976} (denoted here by index $DS$), one gets:
\begin{align*}
m_{DS}(\emptyset)& = 0\\
m_{DS}(\theta_1)&=0.21/[1- k_{12}]=0.21/[1-0.65]=0.21/0.35=0.600000\\
m_{DS}(\theta_2)&=0.11/[1-k_{12}]=0.11/[1-0.65]=0.11/0.35=0.314286\\
m_{DS}(\theta_1\cup\theta_2)&=0.03/[1-k_{12}]=0.03/[1-0.65]=0.03/0.35=0.085714
\end{align*}
\item If one applies {\bf{Smets' rule}} \cite{Smets_1994,Smets_2000} (i.e. the non normalized version of Dempster's rule with the conflicting mass transferred onto the empty set), one gets:
\begin{align*}
m_{S}(\emptyset)&=m(\emptyset)=0.65 \qquad\text{(conflicting mass)}\\
m_{S}(\theta_1)&=0.21\\
m_{S}(\theta_2)&=0.11\\
m_{S}(\theta_1\cup\theta_2)&=0.03
\end{align*}
\end{itemize}
\begin{itemize}
\item If one applies {\bf{Yager's rule}} \cite{Yager_1983,Yager_1985,Yager_1987}, one gets:
\begin{align*}
m_{Y}(\emptyset)&= 0\\
m_{Y}(\theta_1)&=0.21\\
m_{Y}(\theta_2)&=0.11\\
m_{Y}(\theta_1\cup\theta_2)&=0.03 + k_{12}=0.03+0.65=0.68
\end{align*}
\item If one applies  {\bf{Dubois \& Prade's rule}} \cite{Dubois_1988}, one gets because $\theta_3\overset{\mathcal{M}}{=}\emptyset$ :
\begin{align*}
m_{DP}(\emptyset)& = 0 \qquad \text{(by definition of Dubois \& Prade's rule)}\\
m_{DP}(\theta_1)&= [m_1(\theta_1)m_2(\theta_1) + m_1(\theta_1)m_2(\theta_1\cup\theta_2)+ m_2(\theta_1)m_1(\theta_1\cup\theta_2)]\\
& \quad + [m_1(\theta_1)m_2(\theta_3) + m_2(\theta_1)m_1(\theta_3)]\\
& = [0.1\cdot 0.5+0.1\cdot 0.1 +0.5\cdot 0.3] + [0.1\cdot 0.3 +0.5\cdot 0.2] = 0.21 + 0.13 = 0.34\\
m_{DP}(\theta_2)&=[0.4\cdot 0.1 + 0.4\cdot 0.1 +0.1\cdot 0.3] + [0.4\cdot 0.3 + 0.1\cdot 0.2] = 0.11 + 0.14 = 0.25\\
m_{DP}(\theta_1\cup\theta_2)&=[m_1(\theta_1\cup\theta_2)m_2(\theta_1\cup\theta_2)] 
+ [m_1(\theta_1\cup\theta_2)m_2(\theta_3) + m_2(\theta_1\cup\theta_2)m_1(\theta_3)]\\
& \quad + [m_1(\theta_1)m_2(\theta_2) + m_2(\theta_1)m_1(\theta_2)]\\
& = [0.3 0.1 ] + [0.3  \cdot 0.3 + 0.1  \cdot 0.2 ] + [0.1 \cdot 0.1 + 0.5  \cdot 0.4] = [0.03] + [0.09+0.02] + [0.01 + 0.20]\\
& = 0.03 + 0.11 + 0.21 = 0.35
\end{align*}
Now if one adds up the masses, one gets $0+ 0.34+0.25+0.35=0.94$ which is less than 1. Therefore Dubois \& Prade's rule of combination does not work when a singleton, or an union of singletons, becomes empty (in a dynamic fusion problem). The products of such empty-element columns of the mass matrix $\mathbf{M}$ are lost; this problem is fixed in DSmT by the sum $S_2(.)$ in \eqref{eq:DSmHkBis1} which transfers these products to the total or partial ignorances.\\
\end{itemize}
In this particular example, using the hybrid DSm rule, one transfers the product of the empty-element $\theta_3$ column, $m_1(\theta_3)m_2(\theta_3)=0.2\cdot 0.3=0.06$, to $m_{DSmh}(\theta_1\cup\theta_2)$, which becomes equal to $0.35+0.06=0.41$.\\

\subsection{Fusion of imprecise beliefs}
\label{sec2.7}

In many fusion problems, it seems very difficult (if not impossible) to have precise sources of evidence generating precise basic belief assignments (especially when belief functions are provided by human experts), and a more flexible plausible and paradoxical  theory supporting imprecise information becomes necessary. In the previous sections, we presented the fusion of {\it{precise}} uncertain and conflicting/paradoxical generalized basic belief assignments (gbba) in the DSmT framework. We mean here by precise gbba, basic belief functions/masses $m(.)$ defined precisely on the hyper-power set  $D^\Theta$ where each mass $m(X)$, where $X$ belongs to $D^\Theta$, is represented by only one real number belonging to $[0,1]$ such that $\sum_{X\in D^\Theta}m(X)=1$. In this section, we present the DSm fusion rule for dealing with {\it{admissible imprecise generalized basic belief assignments}} $m^I(.)$ defined as real subunitary intervals of $[0,1]$, or even more general as real subunitary sets [i.e. 
sets, not necessarily intervals]. An imprecise belief assignment $m^I(.)$ over $D^\Theta$ is said admissible if and only if there exists for every $X\in D^\Theta$ at least one real number $m(X)\in m^I(X)$ such that $\sum_{X\in D^\Theta}m(X)=1$. The idea to work with imprecise belief structures represented by real subset intervals of $[0,1]$ is not new and has been investigated in \cite{Lamata_1994,Denoeux_1997,Denoeux_1999} and references therein. The proposed works available in the literature, upon our knowledge were limited only to sub-unitary interval combination in the framework of Transferable Belief Model (TBM) developed by Smets \cite{Smets_1994,Smets_2000}. We extend the approach of Lamata \& Moral and Den\oe ux based on subunitary interval-valued masses to subunitary set-valued masses; therefore the 
closed intervals used by Den\oe ux to denote imprecise masses are generalized to any sets included in [0,1], i.e. in our case these sets can be unions of (closed, open, or half-open/half-closed) intervals and/or scalars all in $[0,1]$. Here, the proposed extension is done in the context of the DSmT framework, although it can also apply directly to fusion of imprecise belief structures within TBM as well if the user prefers to adopt TBM rather than DSmT.\\

Before presenting the general formula for the combination of generalized imprecise belief structures, we remind the following set operators involved in the formula. Several numerical examples are given in \cite{DSmTBook_2004a}.

\begin{itemize}
 \item
 {\bf{Addition of sets}}
 \begin{equation*}
 S_{1}\boxplus S_{2} =S_{2}\boxplus S_{1}\triangleq \{ x \mid x = s_{1}+s_{2},  s_{1} \in 
S_{1},s_{2} \in S_{2} \}
 \quad\text{with} \quad 
 \begin{cases}
\inf(S_{1}\boxplus S_{2})=\inf(S_1) + \inf(S_2)\\
\sup(S_{1}\boxplus S_{2})=\sup (S_1) + \sup(S_2)
\end{cases}
\label{eq:addition}
\end{equation*}
\item
{\bf{Subtraction of sets}}
 \begin{equation*}
 S_{1}\boxminus S_{2} \triangleq \{ x \mid x = s_{1}-s_{2},  s_{1} \in 
S_{1}, s_{2} \in S_{2} \}
 \quad\text{with} \quad 
\begin{cases}\inf(S_{1}\boxminus S_{2})=\inf(S_1) - \sup(S_2)\\
\sup(S_{1}\boxminus S_{2})=\sup(S_1) - \inf(S_2)
\end{cases}
\label{eq:addition}
 \end{equation*}
  \item
 {\bf{Multiplication of sets}}
 \begin{equation*}
S_{1}\boxdot S_{2} \triangleq \{ x \mid x = s_{1}\cdot s_{2}, s_{1} \in 
S_{1},s_{2} \in S_{2} \}
 \quad\text{with} \quad 
\begin{cases}
\inf(S_{1}\boxdot S_{2})=\inf(S_{1})\cdot \inf(S_{2})\\
\sup(S_{1}\boxdot S_{2})=\sup(S_{1})\cdot \sup (S_{2})
\end{cases}
\label{eq:multiplication}
 \end{equation*}
\end{itemize}

\subsubsection{DSm rule of combination for imprecise beliefs}
We present the generalization of the DSm rules to combine any type of imprecise belief assignment which may be represented by the union of several sub-unitary (half-) open intervals, (half-)closed intervals and/or sets of points belonging to [0,1]. Several numerical examples are also given. In the sequel, one uses the notation $(a,b)$ for an open interval, $[a,b]$ for a closed interval, and $(a,b]$ or $[a,b)$ for a half open and half closed interval. From the previous operators on sets, one can generalize the DSm rules (classic and hybrid) from scalars to sets in the following way \cite{DSmTBook_2004a} (chap. 6): $\forall A\neq\emptyset \in D^\Theta$,
\begin{equation}
m^I(A) = \underset{\underset{(X_1\cap X_2\cap\ldots\cap X_k)=A}{X_1,X_2,\ldots,X_k\in D^\Theta}}{\boxed{\sum}}
\underset{i=1,\ldots,k}{\boxed{\prod}} m_i^I(X_i)
\label{eq:DSMruleSetsImprecise}
\end{equation}
\noindent
where $\boxed{\sum}$ and $\boxed{\prod}$ represent the summation, and respectively product, of sets.\\

Similarly, one can generalize the hybrid DSm rule from scalars to sets in the following way:
\begin{equation}
m_{\mathcal{M}(\Theta)}^I(A)\triangleq 
\phi(A)\boxdot \Bigl[ S_1^I(A) \boxplus S_2^I(A) \boxplus S_3^I(A)\Bigr]
 \label{eq:DSmHkBisImprecise}
\end{equation}
\noindent
where all sets involved in formulas are in the canonical form and $\phi(A)$ is the {\it{characteristic non emptiness function}} of the set $A$ and $S_1^I(A)$, $S_2^I(A)$ and $S_3^I(A)$ are defined by
\begin{equation}
S_1^I(A)\triangleq
\underset{\underset{X_1\cap X_2\cap\ldots\cap X_k=A}{X_1,X_2,\ldots,X_k\in D^\Theta}}{\boxed{\sum}}
\underset{i=1,\ldots,k}{\boxed{\prod}} m_i^I(X_i)
\label{eq:S1I}
\end{equation}
\begin{equation}
S_2^I(A)\triangleq 
\underset{\underset{[\mathcal{U}=A]\vee [(\mathcal{U}\in\boldsymbol{\emptyset}) \wedge (A=I_t)]}{X_1,X_2,\ldots,X_k\in\boldsymbol{\emptyset}}}{\boxed{\sum}}
\underset{i=1,\ldots,k}{\boxed{\prod}} m_i^I(X_i)
\label{eq:S2I}
\end{equation}
\begin{equation}
S_3^I(A)\triangleq
\underset{\underset{X_1\cap X_2\cap\ldots\cap X_k\in\boldsymbol{\emptyset} }{\underset{X_1\cup X_2\cup\ldots\cup X_k=A}{X_1,X_2,\ldots,X_k\in D^\Theta}}}{\boxed{\sum}}
\underset{i=1,\ldots,k}{\boxed{\prod}} m_i^I(X_i)
\label{eq:S3I}
\end{equation}
In the case when all sets are reduced to points (numbers), the set operations become normal operations with numbers; the sets operations are generalizations of numerical operations. When imprecise belief structures reduce to precise belief structure, DSm rules  \eqref{eq:DSMruleSetsImprecise} and \eqref{eq:DSmHkBisImprecise} reduce to their precise version  \eqref{JDZT}  and  \eqref{eq:DSmHkBis1} respectively.

\subsubsection{Example}

Here is a simple example of fusion with with multiple-interval masses. For simplicity, this example is a particular case when the theorem of admissibility (see \cite{DSmTBook_2004a} p. 138 for details) is verified by a few points, which happen to be just on the bounders. It is an extreme example, because we tried to comprise all kinds of possibilities which may occur in the imprecise or very imprecise fusion. So, let's consider a fusion problem over $\Theta=\{\theta_1,\theta_2\}$, two independent sources of information with the following imprecise admissible belief assignments
\begin{table}[h]
\begin{equation*}
\begin{array}{|c|c|c|}
\hline
A\in D^\Theta & m_1^I(A) & m_2^I(A) \\
\hline
\theta_1 & [0.1,0.2] \cup \{0.3\} & [0.4,0.5]\\
\theta_2 &(0.4,0.6)\cup [0.7,0.8] &  [0,0.4]\cup \{0.5,0.6\}\\
\hline
\end{array}
\end{equation*}
\caption{Inputs of the fusion with imprecise bba}
\label{mytablex1}
\end{table}

\noindent
Using the DSm classic rule for sets, one gets
\begin{equation*}
m^I(\theta_1)=([0.1,0.2] \cup \{0.3\})\boxdot [0.4,0.5] = ([0.1,0.2] \boxdot [0.4,0.5])\cup (\{0.3\}\boxdot [0.4,0.5] )= [0.04,0.10] \cup [0.12,0.15]
\end{equation*}
\begin{align*}
m^I(\theta_2)& =((0.4,0.6)\cup [0.7,0.8] )\boxdot ([0,0.4]\cup \{0.5,0.6\})\\
&= ((0.4,0.6)\boxdot [0,0.4])\cup ((0.4,0.6)\boxdot  \{0.5,0.6\}) \cup ([0.7,0.8]\boxdot [0,0.4]) \cup ([0.7,0.8]\boxdot \{0.5,0.6\})\\
&= (0,0.24)\cup (0.20,0.30) \cup (0.24,0.36)\cup [0,0.32] \cup [0.35,0.40] \cup [0.42,0.48] = [0,0.40] \cup [0.42,0.48]
\end{align*}
\begin{align*}
m^I(\theta_1\cap \theta_2)&=
[([0.1,0.2] \cup \{0.3\})\boxdot([0,0.4]\cup \{0.5,0.6\})] \boxplus [[0.4,0.5]\boxdot ((0.4,0.6)\cup [0.7,0.8]) ]\\
&=[ ([0.1,0.2]\boxdot [0,0.4]) \cup ([0.1,0.2]\boxdot \{0.5,0.6\}) \cup ( \{0.3\}\boxdot [0,0.4]) \cup ( \{0.3\}\boxdot  \{0.5,0.6\})] \\
& \quad \boxplus [ ([0.4,0.5]\boxdot (0.4,0.6)) \cup ([0.4,0.5]\boxdot [0.7,0.8] ) ] \\
& = [[0,0.08]\cup [0.05,0.10]\cup [0.06,0.12] \cup [0,0.12] \cup \{0.15,0.18\}] \boxplus [(0.16,0.30)\cup[0.28,0.40]]\\
&= [[0,0.12]\cup \{0.15,0.18\}]\boxplus (0.16,0.40] =(0.16,0.52] \cup (0.31,0.55] \cup (0.34,0.58]=(0.16,0.58]
\end{align*}
\noindent
Hence finally the fusion admissible result is given by:
\begin{table}[h]
\begin{equation*}
\begin{array}{|c|c|}
\hline
A\in D^\Theta & m^I(A)= [m_1^I \oplus m_2^I](A) \\
\hline
\theta_1 & [0.04,0.10] \cup [0.12,0.15]\\
\theta_2 & [0,0.40] \cup [0.42,0.48] \\
\theta_1\cap \theta_2  & (0.16,0.58]\\
\theta_1\cup \theta_2 & 0 \\
\hline
\end{array}
\end{equation*}
\caption{Fusion result with the DSm classic rule}
\label{mytablex2}
\end{table}

\noindent
If one finds out\footnote{We consider now a dynamic fusion problem.} that $\theta_1\cap \theta_2 \overset{\mathcal{M}}{\equiv}\emptyset$ (this is our hybrid model $\mathcal{M}$ one wants to deal with), then one uses the hybrid DSm rule for sets \eqref{eq:DSmHkBisImprecise}: $m_{\mathcal{M}}^I(\theta_1\cap \theta_2)=0$ and $m_{\mathcal{M}}^I(\theta_1\cup \theta_2)= (0.16,0.58]$, the others imprecise masses are not changed. In other words, one gets now with hybrid DSm rule applied to imprecise beliefs:

\begin{table}[h]
\begin{equation*}
\begin{array}{|c|c|}
\hline
A\in D^\Theta & m_{\mathcal{M}}^I(A)= [m_1^I \oplus m_2^I](A) \\
\hline
\theta_1 & [0.04,0.10] \cup [0.12,0.15]\\
\theta_2 & [0,0.40] \cup [0.42,0.48] \\
\theta_1\cap \theta_2\overset{\mathcal{M}}{\equiv}\emptyset  & 0 \\
\theta_1\cup \theta_2 & (0.16,0.58]\\
\hline
\end{array}
\end{equation*}
\caption{Fusion result with the hybrid DSm rule for $\mathcal{M}$ }
\label{mytablex3}
\end{table}

Let's check now the admissibility conditions and theorem. For the source 1, there exist the precise masses $(m_1(\theta_1)=0.3) \in ([0.1,0.2] \cup \{0.3\})$ and $(m_1(\theta_2)=0.7) \in ((0.4,0.6)\cup [0.7,0.8])$ such that $0.3+0.7=1$. For the source 2, there exist the precise masses $(m_1(\theta_1)=0.4) \in ([0.4,0.5])$ and $(m_2(\theta_2)=0.6) \in ([0,0.4]\cup \{0.5,0.6\})$ such that $0.4+0.6=1$. Therefore both sources associated with $m_1^I(.)$ and $m_2^I(.)$ are admissible imprecise sources of information.\\

It can be easily checked that the DSm classic fusion of $m_1(.)$ and $m_2(.)$ yields the paradoxical basic belief assignment $m(\theta_1)=[m_1\oplus m_2](\theta_1)=0.12$, $m(\theta_2)=[m_1\oplus m_2](\theta_2)=0.42$ and $m(\theta_1\cap \theta_2)=[m_1\oplus m_2](\theta_1\cap \theta_2)=0.46$.
One sees that the admissibility theorem is satisfied since $(m(\theta_1)=0.12)\in (m^I(\theta_1)=[0.04,0.10] \cup [0.12,0.15])$, $(m(\theta_2)=0.42)\in (m^I(\theta_2)=[0,0.40] \cup [0.42,0.48])$ and $(m(\theta_1\cap \theta_2)=0.46)\in (m^I(\theta_1\cap \theta_2)=(0.16,0.58])$ such that $0.12+0.42+0.46=1$. Similarly if one finds out that $\theta_1\cap\theta_2=\emptyset$, then one uses the hybrid DSm rule and one gets: $m(\theta_1\cap\theta_2)=0$ and $m(\theta_1\cup\theta_2)=0.46$; the others remain unchanged. The admissibility theorem still holds, because one can pick at least one number in each subset $m^I(.)$ such that the sum of these numbers is  1. This approach can be also used in the similar manner to obtain imprecise pignistic probabilities from $m^I(.)$ for decision-making under uncertain, paradoxical and imprecise sources of information as well. The generalized pignistic transformation (GPT) is presented in next section.

\section{Proportional Conflict Redistribution rule}

Instead of applying a direct transfer of partial conflicts onto partial uncertainties as with DSmH, the idea behind the Proportional Conflict Redistribution (PCR) rule \cite{Smarandache_2005c,Book_2006} is to transfer (total or partial) conflicting masses to non-empty sets involved in the conflicts proportionally with respect to the masses assigned to them by sources as follows:
\begin{enumerate}
\item calculation the conjunctive rule of the belief masses of sources;
\item calculation the total or partial conflicting masses;
\item redistribution of the (total or partial) conflicting masses to the non-empty sets involved in the conflicts proportionally with respect to their masses assigned by the sources.
\end{enumerate}
The way the conflicting mass is redistributed yields actually several versions of PCR rules. These PCR fusion rules work for any degree of conflict, for any DSm models (Shafer's model, free DSm model or any hybrid DSm model) and both in DST and DSmT frameworks for static or dynamical fusion situations. We present below only the most sophisticated proportional conflict redistribution rule (corresponding to PCR5 in \cite{Smarandache_2005c,Book_2006} but denoted here just PCR for simplicity) since this PCR rule is what we feel the most efficient PCR fusion rule developed so far\footnote{A more intuitive PCR6 rule for the fusion of $s>2$ sources have been proposed by Martin and Osswald in \cite{Book_2006}. PCR5 and PCR6 coincide for $s=2$.}. PCR rule redistributes the partial conflicting mass to the elements involved in the partial conflict, considering the conjunctive normal form of the partial conflict. PCR is what we think the most mathematically exact redistribution of conflicting mass to non-empty sets following the logic of the conjunctive rule. PCR does a better redistribution of the conflicting mass than Dempster's rule sice PCR goes backwards on the tracks of the conjunctive rule and redistributes the conflicting mass only to the sets involved in the conflict and proportionally to their masses put in the conflict. PCR rule is quasi-associative and preserves the neutral impact of the vacuous belief assignment because in any partial conflict, as well in the total conflict (which is a sum of all partial conflicts), the conjunctive normal form of each partial conflict does not include $\Theta$ since $\Theta$ is a neutral element for intersection (conflict), therefore $\Theta$ gets no mass after the redistribution of the conflicting mass. We have also proved in \cite{Book_2006} the continuity property of the PCR result with continuous variations of bba to combine. The general PCR formula for $s\geq 2$ sources is given by \cite{Book_2006} $m_{PCR}(\emptyset)=0$ and $\forall X\in G\setminus\{\emptyset\}$ 
\begin{multline}
m_{PCR}(X) = m_{12\ldots s}(X)+
\sum_{\substack{2\leq t\leq s\\ 1\leq r_1,\ldots,r_t\leq s
1\leq r_1< r_2 < \ldots < r_{t-1} < (r_t=s)}}
 \sum_{\substack{X_{j_2},\ldots,X_{j_t}\in G\setminus\{X\}\\
\{j_2,\ldots,j_t\} \in \mathcal{P}^{t-1}(\{1,\ldots,n\})\\
X\cap X_{j_2}\cap \ldots\cap X_{j_s}=\emptyset\\
\{i_1,\ldots,i_s\} \in \mathcal{P}^{s}(\{1,\ldots,s\})}}\\
\frac{(\prod_{k_1=1}^{r_1} m_{i_{k_1}}(X)^2)\cdot [\prod_{l=2}^{t}( \prod_{k_l=r_{l-1} +1}^{r_l} m_{i_{k_l}}(X_{j_l})]}{(\prod_{k_1=1}^{r_1} m_{i_{k_1}}(X)) + [\sum_{l=2}^{t}( \prod_{k_l=r_{l-1} +1}^{r_l} m_{i_{k_l}}(X_{j_l})]}
\label{eq:PCR5s}
\end{multline}

\noindent
where  all sets involved in formulas are in canonical form (i.e. conjunctive normal form) and where $G$ corresponds to classical power-set $2^\Theta$ if Shafer's model is used or $G$ corresponds to a constrained hyper-power set $D^\Theta$ if any other hybrid DSm model is used instead; $i$, $j$, $k$, $r$, $s$ and $t$ in \eqref{eq:PCR5s} are integers. $m_{12\ldots s}(X)\equiv m_{\cap}(X)$ corresponds to the conjunctive consensus on $X$ between $s$ sources and where all denominators are different from zero. If a denominator is zero, that fraction is discarded; the set of all subsets of $k$ elements from $\{1,2,\ldots,n\}$ (permutations of $n$ elements taken by $k$) was denoted  $\mathcal{P}^{k}(\{1,2,\ldots,n\})$, the order of elements doesn't count. \\

When $s=2$ (fusion of only two sources), the previous PCR rule reduces to its simple following fusion formula: $m_{PCR}(\emptyset)=0$ and $\forall X\in G\setminus\{\emptyset\}$
\begin{equation}
m_{PCR}(X)=m_{12}(X) +
\sum_{\substack{Y\in G\setminus\{X\} \\ X\cap Y=\emptyset}} 
[\frac{m_1(X)^2m_2(Y)}{m_1(X)+m_2(Y)} +
 \frac{m_2(X)^2 m_1(Y)}{m_2(X)+m_1(Y)}]
   \label{eq:PCR5}
 \end{equation}
\noindent

\subsection{Examples}

\begin{itemize}
\item {\bf{Example 1}}: Let's take $\Theta=\{A, B\}$ of exclusive elements (Shafer's model), and the following bba:
\begin{center}
\begin{tabular}[h]{|c|ccc|}
\hline
 & $A$ & $B$ & $ A\cup B$\\
 \hline
 $m_1(.)$ & 0.6 & 0 & 0.4 \\
 \hline
 $m_2(.)$ & 0 & 0.3 & 0.7 \\
 \hline
 \hline
 $m_{\cap}(.)$ & 0.42 & 0.12 & 0.28 \\
 \hline
\end{tabular}
\end{center}
The conflicting mass is $k_{12}=m_{\cap}(A\cap B)$ and equals $m_1(A)m_2(B)+m_1(B)m_2(A)=0.18$.
Therefore $A$ and $B$ are the only focal elements involved in the conflict. Hence according to the PCR hypothesis only $A$ and $B$ deserve a part of the conflicting mass and $A\cup B$ do not deserve. With PCR, one redistributes the conflicting mass $k_{12}=0.18$ to $A$ and $B$ proportionally with the masses $m_1(A)$
and $m_2(B)$ assigned to $A$ and $B$ respectively. Here are the results obtained from Dempster's rule, DSmH and PCR:
\begin{center}
\begin{tabular}[h]{|l||ccc|}
\hline
  & $A$ & $B$ & $A\cup B$\\
 \hline
  $m_{DS}$ & 0.512 & 0.146 & 0.342 \\
  $m_{DSmH}$ & 0.420 & 0.120 & 0.460 \\
  $m_{PCR}$ & 0.540 & 0.180 &  0.280 \\
\hline
\end{tabular}
\end{center}
\noindent

\item {\bf{Example 2}}: Let's modify example 1 and consider
\begin{center}
\begin{tabular}[h]{|c|ccc|}
\hline
 & $A$ & $B$ & $ A\cup B$\\
 \hline
 $m_1(.)$ & 0.6 & 0 & 0.4 \\
 \hline
 $m_2(.)$ & 0.2 & 0.3 & 0.5 \\
 \hline
 \hline
 $m_{\cap}(.)$ & 0.50 & 0.12 & 0.20 \\
 \hline
\end{tabular}
\end{center}
\noindent The conflicting mass $k_{12}=m_{\cap}(A\cap B)$ as well as the distribution coefficients for the PCR remains the same as in the previous example but one gets now
\begin{center}
\begin{tabular}[h]{|l||ccc|}
\hline
  & $A$ & $B$ & $A\cup B$\\
 \hline
  $m_{DS}$ & 0.609 & 0.146 & 0.231 \\
 $m_{DSmH}$ & 0.500 & 0.120 & 0.380 \\
 $m_{PCR}$ & 0.620 & 0.180 &  0.200 \\
\hline
\end{tabular}
\end{center}

\item {\bf{Example 3}}: Let's modify example 2 and consider
\begin{center}
\begin{tabular}[h]{|c|ccc|}
\hline
 & $A$ & $B$ & $ A\cup B$\\
 \hline
 $m_1(.)$ & 0.6 & 0.3 & 0.1 \\
 \hline
 $m_2(.)$ & 0.2 & 0.3 & 0.5 \\
 \hline
 \hline
 $m_{\cap}(.)$ & 0.44 & 0.27 & 0.05 \\
 \hline
\end{tabular}
\end{center}
The conflicting mass $k_{12}=0.24 = m_1(A)m_2(B)+m_1(B)m_2(A)=0.24$ is now different from previous examples, which means that $m_2(A) = 0.2$ and $m_1(B) =0.3$ did make an impact on the conflict. Therefore $A$ and $B$ are the only focal elements involved in the conflict and thus only $A$ and $B$ deserve a part of the conflicting mass. PCR redistributes the partial conflicting mass 0.18 to $A$ and $B$ proportionally with the masses $m_1(A)$ and $m_2(B)$ and also the partial conflicting mass 0.06 to $A$ and $B$ proportionally with the masses $m_2(A)$ and $m_1(B)$. After all derivations (see \cite{Florea_2006} for details), one finally gets
\begin{center}
\begin{tabular}[h]{|l||ccc|}
\hline
  & $A$ & $B$ & $A\cup B$\\
 \hline
  $m_{DS}$ & 0.579 & 0.355 & 0.066 \\
 $m_{DSmH}$ & 0.440 & 0.270 & 0.290 \\
 $m_{PCR}$ & 0.584 & 0.366 &  0.050 \\
\hline
\end{tabular}
\end{center}
\noindent

One clearly sees that $m_{DS}(A\cup B)$ gets some mass from the conflicting mass although $A\cup B$ does not deserve any part of the conflicting mass (according to PCR hypothesis) since $A\cup B$ is not involved in the conflict (only $A$ and $B$ are involved in the conflicting mass). Dempster's rule appears to us less exact than PCR and Inagaki's rules \cite{Inagaki_1991}. It can be showed  \cite{Florea_2006} that Inagaki's fusion rule (with an optimal choice of tuning parameters) can become in some cases very close to  PCR but upon our opinion PCR result is more exact (at least less ad-hoc than Inagaki's one).
\end{itemize}

\subsection{Zadeh's example}

We compare here the solutions for well-known Zadeh's example \cite{Zadeh_1979,Zadeh_1986} provided by several fusion rules. A detailed presentation with more comparisons can be found in \cite{DSmTBook_2004a,Book_2006}. Let's consider $\Theta=\{M,C,T\}$ as the frame of three potential origins about possible diseases of a patient ($M$ standing for {\it{meningitis}}, $C$ for {\it{concussion}} and $T$ for {\it{tumor}}), the Shafer's model and the two following belief assignments provided by two independent doctors after examination of the same patient.
\begin{align*}
m_1(M)&=0.9 &\quad m_1(C)&=0 &\quad m_1(T)&=0.1\\
m_2(M)&=0 &\quad m_2(C)&=0.9 &\quad m_2(T)&=0.1
\end{align*}
The total conflicting mass is high since it is
$$m_1(M)m_2(C)+m_1(M)m_2(T)+m_2(C)m_1(T)=0.99$$
\begin{itemize}
\item with Dempster's rule and Shafer's model (DS), one gets the counter-intuitive result (see justifications in \cite{Zadeh_1979,Dubois_1986c,Yager_1987,Voorbraak_1991,DSmTBook_2004a}): $m_{DS}(T)=1$
\item with Yager's rule \cite{Yager_1987} and Shafer's model: 
$m_{Y}(M\cup C \cup T)=0.99$ and $m_{Y}(T)=0.01$
\item with DSmH and Shafer's model:
$$m_{DSmH}(M\cup C)=0.81\qquad  m_{DSmH}(T)=0.01\qquad m_{DSmH}(M\cup T)=m_{DSmH}(C\cup T)=0.09$$
\item The Dubois \& Prade's rule (DP) \cite{Dubois_1986c} based on Shafer's model provides in Zadeh's example the same result as DSmH, because DP and DSmH coincide in all static fusion problems\footnote{Indeed DP rule has been developed for static fusion only while DSmH has been developed to take into account the possible dynamicity of the frame itself and also its associated model.}.
\item with PCR and Shafer's model:
$$m_{PCR}(M)=m_{PCR}(C)=0.486/qquad  m_{PCR}(T)=0.028$$
\end{itemize}

One sees that when the total conflict between sources becomes high, DSmT is able (upon authors opinion) to manage more adequately through DSmH or PCR rules the combination of information than Dempster's rule, even when working with Shafer's model - which is only a specific hybrid model. DSmH rule is in agreement with DP rule for the static fusion, but DSmH and DP rules differ in general (for non degenerate cases) for dynamic fusion while PCR rule is the most exact proportional conflict redistribution rule. Besides this particular example, we showed in \cite{DSmTBook_2004a} that there exist several infinite classes of counter-examples to Dempster's rule which can be solved by DSmT. \\

In summary, DST based on Dempster's rule  provides counter-intutive results in Zadeh's example, or in non-Bayesian examples similar to Zadeh's and no result when the conflict is 1. Only ad-hoc discounting techniques allow to circumvent troubles of Dempster's rule or we need to switch to another model of representation/frame; in the later case the solution obtained doesn't fit with the Shafer's model one originally wanted to work with. We want also to emphasize that in dynamic fusion when the conflict becomes high, both DST \cite{Shafer_1976} and Smets' Transferable Belief Model (TBM) \cite{Smets_1994} approaches fail to respond to new information provided by new sources. This can be easily showed by the very simple following example. Let's consider $\Theta=\{A,B,C\}$ and the following (precise) belief assignments $m_1(A)=0.4$, $m_1(C)=0.6$ and $m_2(A)=0.7$, $m_2(B)=0.3$. Then one gets\footnote{We introduce here explicitly the indexes of sources in the fusion result since more than two sources are considered in this example.} with Dempster's rule, Smets' TBM (i.e. the non-normalized version of Dempster's combination), (DSmH) and (PCR5): $m_{DS}^{12}(A)=1$, $m_{TBM}^{12}(A)=0.28$, $m_{TBM}^{12}(\emptyset)=0.72$, 
\begin{equation*}
\begin{cases}
m_{DSmH}^{12}(A)=0.28\\
m_{DSmH}^{12}(A\cup B)=0.12\\
m_{DSmH}^{12}(A\cup C)=0.42\\
m_{DSmH}^{12}(B\cup C)=0.18
\end{cases}
\end{equation*}
\begin{equation*}
\begin{cases}
m_{PCR}^{12}(A)=0.574725\\
m_{PCR}^{12}(B)=0.111429\\
m_{PCR}^{12}(C)=0.313846
\end{cases}
\end{equation*}

Now let's consider a temporal fusion problem and introduce a third source $m_3(.)$ with $m_3(B)=0.8$ and $m_3(C)=0.2$. Then one sequentially combines the results obtained by $m_{TBM}^{12}(.)$, $m_{DS}^{12}(.)$, $m_{DSmH}^{12}(.)$ and $m_{PCR}^{12}(.)$ with
the new evidence $m_3(.)$ and one sees that $m_{DS}^{(12)3}$ becomes not defined (division by zero) and $m_{TBM}^{(12)3}(\emptyset)=1$ while (DSmH) and (PCR) provide 
\begin{equation*}
\begin{cases}
m_{DSmH}^{(12)3}(B)=0.240\\
m_{DSmH}^{(12)3}(C)=0.120\\
m_{DSmH}^{(12)3}(A\cup B)=0.224\\
m_{DSmH}^{(12)3}(A\cup C)= 0.056\\
m_{DSmH}^{(12)3}(A\cup B\cup C)= 0.360\\
\end{cases}
\end{equation*}
\begin{equation*}
\begin{cases}
m_{PCR}^{(12)3}(A)=0.277490\\
m_{PCR}^{(12)3}(B)=0.545010\\
m_{PCR}^{(12)3}(C)=0.177500
\end{cases}
\end{equation*}

When the mass committed to empty set becomes one at a previous temporal fusion step, then both DST and TBM do not respond to new information. Let's continue the example and consider a fourth source $m_4(.)$ with $m_4(A)=0.5$, $m_4(B)=0.3$ and $m_4(C)=0.2$. Then it is easy to see that $m_{DS}^{((12)3)4}(.)$ is not defined since at previous step $m_{DS}^{(12)3}(.)$ was already not defined, and that $m_{TBM}^{((12)3)4}(\emptyset)=1$ whatever $m_4(.)$ is because at the previous fusion step one had $m_{TBM}^{(12)3}(\emptyset)=1$. Therefore for a number of sources $n\geq 2$, DST and TBM approaches do not respond to new information incoming in the fusion process while both (DSmH) and (PCR) rules respond to new information. To make DST and/or TBM working properly in such cases, it is necessary to introduce ad-hoc temporal discounting techniques which are not necessary to introduce if DSmT is adopted. If there are good reasons to introduce temporal discounting, there is obviously no difficulty to apply the DSm fusion of these discounted sources. A analysis of this behavior for target type tracking is presented in \cite{Dezert_2006,Book_2006}.

\subsection{The generalized pignistic transformation (GPT)}

\subsubsection{The classical pignistic transformation}

We follow here the Smets'  vision which considers the management of information as a two 2-levels process: credal (for combination of evidences) and pignistic\footnote{Pignistic terminology has been coined by Philippe Smets and comes from {\it{pignus}}, a bet in Latin.} (for decision-making) , i.e "{\it{when someone must take a decision, he must then construct a probability function derived from the belief function that describes his credal state. This probability function is then used to make decisions}}" \cite{Smets_1988} (p. 284). One obvious way to build this probability function corresponds to the so-called Classical Pignistic Transformation (CPT) defined in the DST framework (i.e. based on the Shafer's model assumption) as \cite{Smets_2000}:

\begin{equation}
P\{A\}=\sum_{X \in 2^\Theta}\frac{|X\cap A|}{|X|}m(X)
\label{eq:Pig}
\end{equation}

where $|A|$ denotes the number of worlds in the set $A$ (with convention $|\emptyset | / |\emptyset |=1$, to define $P\{\emptyset \}$). $P\{A\}$ corresponds to $BetP(A)$ in Smets' notation \cite{Smets_2000}. Decisions are achieved by computing the expected utilities of the acts using the subjective/pignistic $P\{.\}$ as the probability function needed to compute expectations.
Usually, one uses the maximum of the pignistic probability as decision criterion. The max. of $P\{.\}$ is often considered as a prudent betting decision criterion between the two other alternatives (max of plausibility or max. of credibility which appears to be respectively too optimistic or too pessimistic). It is easy to show that $P\{.\}$ is indeed a probability function (see \cite{Smets_1994}).

\subsubsection{Notion of DSm cardinality}
One important notion involved in the definition of the Generalized Pignistic Transformation (GPT) is the {\it{DSm cardinality}}. The {\it{DSm cardinality}} of any element $A$ of hyper-power set  $D^\Theta$, denoted $\mathcal{C}_\mathcal{M}(A)$, corresponds to the number of parts of $A$ in the corresponding fuzzy/vague Venn diagram of the problem (model $\mathcal{M}$) taking into account the set of integrity constraints (if any), i.e. all the possible intersections due to the nature of the elements $\theta_i$. This {\it{intrinsic cardinality}} depends on the model $\mathcal{M}$ (free, hybrid or Shafer's model).  $\mathcal{M}$ is the model that contains $A$, which depends both on the
dimension $n=\vert \Theta \vert$ and on the number of non-empty intersections present in its associated Venn diagram (see \cite{DSmTBook_2004a} for details ). The DSm cardinality depends on the cardinal of $\Theta = \{\theta_1,\theta_2,\ldots,\theta_n\}$ and on the model of $D^\Theta$ (i.e., the number of intersections and between what elements of $\Theta$ - in a word the structure) at the same time; it is not necessarily that every singleton, say $\theta_i$, has the same DSm cardinal, because each singleton has a different structure; if its structure is the simplest (no intersection of this elements with other elements) then $\mathcal{C}_\mathcal{M}(\theta_i)=1$, if the structure is more complicated (many intersections) then $\mathcal{C}_\mathcal{M}(\theta_i) > 1$; let's consider a singleton $\theta_i$: if it has 1 intersection only then $\mathcal{C}_\mathcal{M}(\theta_i)=2$, for 2 intersections only $\mathcal{C}_\mathcal{M}(\theta_i)$ is 3 or 4 depending on the model $\mathcal{M}$, for $m$ intersections it is between $m+1$ and $2^m$ depending on the model; the maximum DSm cardinality is $2^{n-1}$ and occurs for $\theta_1\cup\theta_2\cup\ldots\cup\theta_n$ in the free model $\mathcal{M}^f$; similarly for any set from $D^\Theta$: the more complicated structure it has, the bigger is the DSm cardinal;
thus the DSm cardinality measures the complexity of en element from $D^\Theta$, which is a nice characterization in our opinion; we may say that for the singleton $\theta_i$ not even $\vert\Theta\vert$ counts, but only its structure (= how many other singletons intersect $\theta_i$). Simple illustrative examples are given in Chapter 3 and 7 of \cite{DSmTBook_2004a}.
One has $1 \leq \mathcal{C}_\mathcal{M}(A) \leq 2^n-1$. $\mathcal{C}_\mathcal{M}(A)$ must not be confused with the classical cardinality $\vert A \vert$ of a given set $A$ (i.e. the number of its distinct elements) - that's why a new notation is necessary here.
$\mathcal{C}_\mathcal{M}(A)$ is very easy to compute by programming from the algorithm of generation of $D^\Theta$ given explicated in \cite{DSmTBook_2004a}.

As example, let's take back the example of the simple hybrid DSm model described in section \ref{Sec:DSMmodels}, then one gets the following list of elements (with their DSm cardinal) for the restricted $D^\Theta$ taking into account the integrity constraints of this hybrid model:
\begin{equation*}
\begin{array}{lcl}
A\in D^\Theta                     & \mathcal{C}_{\mathcal{M}}(A) \\
\hline
\alpha_0\triangleq\emptyset                                                 & 0 \\
\alpha_1\triangleq\theta_1\cap\theta_2                             & 1 \\
\alpha_2\triangleq\theta_3                                                    & 1 \\
\alpha_3\triangleq\theta_1                                                    & 2 \\
\alpha_4\triangleq\theta_2                                                    & 2 \\
\alpha_5\triangleq\theta_1\cup\theta_2                             & 3 \\
\alpha_6\triangleq\theta_1\cup\theta_3                             & 3 \\
\alpha_7\triangleq\theta_2\cup\theta_3                             & 3 \\
\alpha_8\triangleq\theta_1\cup\theta_2\cup\theta_3       & 4  \\
\end{array}
\end{equation*}
\begin{center}
{\bf{Example of DSm cardinals}}: $\mathcal{C}_{\mathcal{M}}(A)$ for hybrid model $\mathcal{M}$
\end{center}

\subsubsection{The  Generalized Pignistic Transformation}

To take a rational decision within the DSmT framework, it is necessary to generalize the Classical Pignistic Transformation in order to construct a pignistic probability function from any generalized basic belief assignment $m(.)$ drawn from the DSm rules of combination. Here is the simplest and direct extension of the CPT to define the Generalized Pignistic Transformation:
\begin{equation}
\forall A \in D^\Theta, \qquad \qquad P\{A\}=\sum_{X \in D^\Theta}  \frac{\mathcal{C}_{\mathcal{M}}(X\cap A)}{\mathcal{C}_{\mathcal{M}}(X)}m(X)
\label{eq:PigG}
\end{equation}
\noindent
where $\mathcal{C}_{\mathcal{M}}(X)$ denotes the DSm cardinal of proposition $X$ for the DSm model $\mathcal{M}$ of the problem under consideration.\\

The decision about the solution of the problem  is usually taken by the maximum of pignistic probability function $P\{.\}$. Let's remark the close ressemblance of the two pignistic transformations \eqref{eq:Pig} and \eqref{eq:PigG}. It can be shown that \eqref{eq:PigG} reduces to \eqref{eq:Pig} when the hyper-power set $D^\Theta$ reduces to classical power set $2^\Theta$ if we adopt Shafer's model. But  \eqref{eq:PigG} is a generalization of  \eqref{eq:Pig} since it can be used for computing pignistic probabilities for any models (including Shafer's model). It has been proved in \cite{DSmTBook_2004a} (Chap. 7) that $P\{.\}$ is indeed a probability function.

\section{Fusion of qualitative beliefs}

We introduce here the notion of qualitative belief assignment to model beliefs of human experts expressed in natural language (with linguistic labels). We show how qualitative beliefs can be efficiently combined using an extension of Dezert-Smarandache Theory (DSmT) of plausible and paradoxical quantitative reasoning to qualitative reasoning shortly presented in previous sections. A more detailed presentation can be found in \cite{Book_2006}. The derivations are based on a new arithmetic on linguistic labels which allows a direct extension of classical DSm fusion rule or DSm Hybrid rules. An approximate qualitative PCR5 rule is also presented.

\subsection{Qualitative Operators}

Computing with words (CW) and qualitative information is more vague, less precise than computing with numbers, but it offers the advantage of robustness if done correctly. Here is a general arithmetic we propose for computing with words (i.e. with linguistic labels). Let's consider a finite frame $\Theta=\{\theta_1,\ldots,\theta_n\}$ of $n$ (exhaustive) elements $\theta_i$, $i=1,2,\ldots,n$, with an associated model $\mathcal{M}(\Theta)$ on $\Theta$ (either Shafer's model $\mathcal{M}^0(\Theta)$, free-DSm model $\mathcal{M}^f(\Theta)$, or more general any Hybrid-DSm model \cite{DSmTBook_2004a}). A model $\mathcal{M}(\Theta)$ is defined by the set of integrity constraints on elements of $\Theta$ (if any); Shafer's model $\mathcal{M}^0(\Theta)$ assumes all elements of $\Theta$ truly exclusive, while free-DSm model $\mathcal{M}^f(\Theta)$ assumes no exclusivity constraints between elements of the frame $\Theta$. Let's define a finite set of linguistic labels\index{linguistic labels} $\tilde{L}=\{L_1,L_2,\ldots,L_m\}$ where $m\geq 2$ is an integer. $\tilde{L}$ is endowed with a total order relationship $\prec$, so that $L_1\prec L_2\prec \ldots\prec L_m$. To work on a close linguistic set under linguistic addition and multiplication operators, we extends $\tilde{L}$ with two extreme values $L_{0}$ and $L_{m+1}$ where $L_{0}$ corresponds to the minimal qualitative value and $L_{m+1}$ corresponds to the maximal qualitative value, in such a way that
$$L_0\prec L_1\prec L_2\prec \ldots\prec L_m\prec L_{m+1}$$
\noindent
where $\prec$ means inferior to, or less (in quality) than, or smaller (in quality) than, etc. hence a relation of order from a qualitative point of view. But if we make a correspondence between qualitative
labels and quantitative values on the scale $[0, 1]$, then $L_{\min}=L_0$ would correspond to the numerical value 0, while $L_{\max}=L_{m+1}$ would correspond to the
numerical value 1, and each $L_i$ would belong to $[0,1]$, i. e.
$$L_{\min}=L_0 < L_1 < L_2 < \ldots <L_m < L_{m+1}=L_{\max}$$

\noindent
From now on, we work on extended ordered set $L$ of qualitative values
$$L=\{L_0,\tilde{L},L_{m+1}\}=\{L_0,L_1,L_2,\ldots,L_m,L_{m+1}\}$$
\noindent
The qualitative addition and multiplication operators are respectively defined in the following way:
\begin{itemize}
\item Addition :
\begin{equation}
L_i + L_j=
\begin{cases}
L_{i+j}, & \text{if}\ i+j < m+1,\\
L_{m+1}, & \text{if}\ i+j \geq m+1.
\end{cases}
\label{qadd}
\end{equation}
\item Multiplication :
\begin{equation}
L_i \times L_j=L_{\min\{i,j\}}
\label{qmult}
\end{equation}
\end{itemize}

\noindent
These two operators are well-defined, commutative, associative, and unitary\index{unitary operator}. Addition of labels is a unitary operation\index{unitary operator} since $L_0 =
L_{\min}$ is the unitary element, i.e. $L_i + L_0 = L_0 + L_i = L_{i+0} = L_i$ for all $0\leq  i \leq
m+1$.
Multiplication of labels is also a unitary operation\index{unitary operator} since $L_{m+1} = L_{\max}$ is the unitary element, i.e.
$L_i \times L_{m+1} = L_{m+1} \times L_i = L_{\min\{i, m+1\}} = L_i$ for $0\leq  i \leq m+1$. $L_0$ is the unit element for addition, while $L_{m+1}$ is the unit element for multiplication. $L$ is closed under $+$ and $\times$. The mathematical structure formed by $(L, +, \times)$ is a commutative bisemigroup\index{bisemigroup} with different unitary elements for each operation. We recall that a bisemigroup\index{bisemigroup} is a set $S$ endowed with two associative binary operations such that $S$ is closed under both operations. If $L$ is not an exhaustive set of qualitative labels, then other labels may exist in between the initial ones, so we can work with labels and numbers - since a
refinement of $L$ is possible. When mapping from $L$ to crisp numbers or intervals, $L_0 = 0$ and $L_{m+1}=1$, while $0<L_i<1$, for all $i$, as crisp numbers, or $L_i$ included in $[0,1]$ as intervals/subsets.\\

For example, $L_1$, $L_2$, $L_3$ and $L_4$ may represent the following qualitative values: $L_1\triangleq \text{very poor}$, $L_2\triangleq \text{poor}$, $L_3\triangleq \text{good}$ and $L_4\triangleq \text{very good}$ where $\triangleq$ symbol means "by definition".\\

We think it is better to define the multiplication $\times$ of $L_i\times L_j$ by $L_{\min\{i,j\}}$ because multiplying two numbers $a$ and $b$ in $[0,1]$ one gets a result which is less than each of them, the product is not bigger than both of them as Bolanos et al. did in \cite{Bolanos_1993} by approximating $L_i\times L_j= L_{i+j} > \max\{L_i, L_j\}$.  While for the addition it is the opposite: adding two numbers in the interval $[0,1]$ the sum should be bigger than both of them, not smaller as in \cite{Bolanos_1993} case where $L_i+L_j = \min\{L_i, L_j\}< \max\{L_i,
L_j\}$.

\subsection{Qualitative Belief Assignment}
A qualitative belief assignment\index{qualitative belief assignment} (qba), and we call it {\it{qualitative belief mass}} or {\it{q-mass}} for short, is a mapping function $qm(.): G^\Theta \mapsto L$ where $G^\Theta$ corresponds the space of propositions generated with $\cap$ and $\cup$ operators and elements of $\Theta$ taking into account the integrity constraints of the model. For example if Shafer's model is chosen for $\Theta$, then $G^\Theta$ is nothing but the classical power set $2^\Theta$ \cite{Shafer_1976}, whereas if free DSm model is adopted $G^\Theta$ will correspond to Dedekind's lattice (hyper-power set) $D^\Theta$ \cite{DSmTBook_2004a}. Note that in this qualitative framework, there is no way to define normalized $qm(.)$, but qualitative quasi-normalization\index{quasi-normalization} is still possible as seen further. Using the qualitative operations defined previously we can easily extend the combination rules from quantitative to qualitative. In the sequel we will consider $s\geq 2$ qualitative belief assignments\index{qualitative belief assignment} $qm_1(.),\ldots, qm_s(.)$ defined over the same space $G^\Theta$ and provided by $s$ independent sources $S_1,\ldots,S_s$ of evidence. \\

 \noindent{\bf{Important note}}: The addition and multiplication operators used in all qualitative fusion formulas in next sections correspond to {\it{qualitative addition}} and {\it{qualitative multiplication}} operators defined in \eqref{qadd} and \eqref{qmult} and must not be confused with classical addition and multiplication operators for numbers.
 
\subsection{Qualitative Conjunctive Rule}

The qualitative Conjunctive Rule (qCR)\index{qualitative Conjunctive Rule (qCR)} of $s\geq 2$ sources is defined similarly to the quantitative conjunctive consensus rule, i.e.

\begin{equation}
qm_{qCR}(X)=\sum_{\substack{X_1,\ldots,X_s\in G^\Theta\\ X_1\cap \ldots \cap X_s=X}} \prod_{i=1}^{s} qm_i(X_i)
\label{qCR}
\end{equation}

\noindent
The total qualitative conflicting mass is given by $$K_{1\ldots s}=\sum_{\substack{X_1,\ldots,X_s\in G^\Theta\\ X_1\cap \ldots \cap X_s=\emptyset}} \prod_{i=1}^{s} qm_i(X_i)$$

\subsection{Qualitative DSm Classic rule}

The qualitative DSm Classic rule (q-DSmC)\index{qualitative DSm Classic rule (q-DSmC)} for $s\geq 2$  is defined similarly to DSm Classic fusion rule (DSmC) as follows : $qm_{qDSmC}(\emptyset)=L_0$ and for all $X\in D^\Theta\setminus \{\emptyset\}$,

\begin{equation}
qm_{qDSmC}(X)=\sum_{\substack{X_1,,\ldots,X_s\in D^\Theta\\ X_1\cap\ldots \cap X_s=X}} \prod_{i=1}^{s} qm_i(X_i)
\label{qDSmC}
\end{equation}

\subsection{Qualitative DSm Hybrid rule}

The qualitative DSm Hybrid rule (q-DSmH)\index{qualitative DSm Hybrid rule (q-DSmH)} is defined similarly to quantitative DSm hybrid rule \cite{DSmTBook_2004a} as follows: 
\begin{equation}
qm_{qDSmH}(\emptyset)=L_0
\end{equation}
\noindent
and for all $X\in G^\Theta\setminus \{\emptyset\}$
\begin{equation}
qm_{qDSmH}(X)\triangleq
\phi(X)\cdot\Bigl[ qS_1(X) + qS_2(X) + qS_3(X)\Bigr]
 \label{qDSmH}
\end{equation}
\noindent
where all sets involved in formulas are in the canonical form and $\phi(X)$ is the {\it{characteristic non-emptiness function}} of a set $X$, i.e. $\phi(X)= L_{m+1}$ if  $X\notin \boldsymbol{\emptyset}$ and $\phi(X)= L_0$ otherwise, where $\boldsymbol{\emptyset}\triangleq\{\boldsymbol{\emptyset}_{\mathcal{M}},\emptyset\}$. $\boldsymbol{\emptyset}_{\mathcal{M}}$ is the set  of all elements of $D^\Theta$ which have been forced to be empty through the constraints of the model $\mathcal{M}$ and $\emptyset$ is the classical/universal empty set. $qS_1(X)\equiv qm_{qDSmC}(X)$, $qS_2(X)$, $qS_3(X)$ are defined by
\begin{equation}
qS_1(X)\triangleq \sum_{\substack{X_1,X_2,\ldots,X_s\in D^\Theta\\ X_1\cap X_2\cap\ldots\cap X_s=X}} \prod_{i=1}^{s} qm_i(X_i)
\end{equation}
\begin{equation}
qS_2(X)\triangleq \sum_{\substack{X_1,X_2,\ldots,X_s\in\boldsymbol{\emptyset}\\ [\mathcal{U}=X]\vee [(\mathcal{U}\in\boldsymbol{\emptyset}) \wedge (X=I_t)]}} \prod_{i=1}^{s} qm_i(X_i)
\end{equation}
\begin{equation}
qS_3(X)\triangleq\sum_{\substack{X_1,X_2,\ldots,X_k\in D^\Theta \\ 
X_1\cup X_2\cup \ldots \cup X_s=X \\ X_1\cap X_2\cap \ldots\cap X_s \in\boldsymbol{\emptyset}}}  \prod_{i=1}^{s} qm_i(X_i)
\end{equation}
\noindent
with $\mathcal{U}\triangleq u(X_1)\cup \ldots \cup u(X_s)$ where $u(X)$ is the union of all $\theta_i$ that compose $X$, $I_t \triangleq \theta_1\cup \ldots\cup \theta_n$ is the total ignorance. $qS_1(X)$ is nothing but the qDSmC rule for $s$ independent sources based on $\mathcal{M}^f(\Theta)$; $qS_2(X)$ is the qualitative mass\index{qualitative mass} of all relatively and absolutely empty sets which is transferred to the total or relative ignorances associated with non existential constraints (if any, like in some dynamic problems); $qS_3(X)$ transfers the sum of relatively empty sets directly onto the canonical disjunctive form of non-empty sets. qDSmH generalizes qDSmC works for any models (free DSm model, Shafer's model or any hybrid models) when manipulating qualitative belief assignments\index{qualitative belief assignment}.

\subsection{Qualitative PCR5 rule (q-PCR5)\index{qualitative PCR5 rule (q-PCR5)}}

In classical (i.e. quantitative) DSmT\index{Dezert-Smarandache Theory (DSmT)} framework, the Proportional Conflict Redistribution rule no.~5 (PCR5) defined in \cite{Book_2006} has been proven to provide very good and coherent results for combining (quantitative) belief masses, see \cite{Smarandache_2005c,Dezert_2006}. When dealing with qualitative beliefs and using Dempster-Shafer Theory (DST)\index{Dempster-Shafer Theory (DST)}, we unfortunately cannot normalize, since it is not possible to divide linguistic labels\index{linguistic labels} by linguistic labels\index{linguistic labels}. Previous authors have used the un-normalized Dempster's rule\index{Dempster's rule}, which actually is equivalent to the Conjunctive Rule in Shafer's model and respectively to DSm conjunctive rule in hybrid and free DSm models. Following the idea of (quantitative) PCR5 fusion rule, we can however use a rough approximation for a qualitative version of PCR5 (denoted qPCR5) as it will be presented in next example, but we did not succeed so far to get a general formula for qualitative PCR5 fusion rule (q-PCR5) because the division of labels could not be defined.

\subsection{A simple example of qualitative Fusion of qba's}

Let's consider the following set of ordered linguistic labels\index{linguistic labels} $L=\{L_0,L_1,L_2,L_3,L_4,L_{5}\}$ (for example, $L_1$, $L_2$, $L_3$ and $L_4$ may represent the values: $L_1\triangleq \text{{\it{very poor}}}$, $L_2\triangleq \text{{\it{poor}}}$, $L_3\triangleq \text{{\it{good}}}$ and $L_4\triangleq \text{{\it{very good}}}$, where $\triangleq$ symbol means  {\it{by definition}}), then addition and multiplication tables are:

\begin{table}[h]
\centering 
\begin{tabular}{|c|cccccc|}
\hline
$+$     & $L_0$ & $L_1$ & $L_2$ & $L_3$ & $L_4$ & $L_5$\\
\hline
$L_0$ & $L_0$ & $L_1$ & $L_2$ & $L_3$ & $L_4$ & $L_5$\\
$L_1$ & $L_1$ & $L_2$ & $L_3$ & $L_4$ & $L_5$ & $L_5$\\
$L_2$ & $L_2$ & $L_3$ & $L_4$ & $L_5$ & $L_5$ & $L_5$\\
$L_3$ & $L_3$ & $L_4$ & $L_5$ & $L_5$ & $L_5$ & $L_5$\\
$L_4$ & $L_4$ & $L_5$ & $L_5$ & $L_5$ & $L_5$ & $L_5$\\
$L_5$ & $L_5$ & $L_5$ & $L_5$ & $L_5$ & $L_5$ & $L_5$\\
\hline
\end{tabular}
\caption{Addition table}
\label{CWTable3}
\end{table}
\begin{table}[h]
\centering 
\begin{tabular}{|c|cccccc|}
\hline
$\times$ & $L_0$ & $L_1$ & $L_2$ & $L_3$ & $L_4$ & $L_5$\\
\hline
$L_0$ & $L_0$ & $L_0$ & $L_0$ & $L_0$ & $L_0$ & $L_0$\\
$L_1$ & $L_0$ & $L_1$ & $L_1$ & $L_1$ & $L_1$ & $L_1$\\
$L_2$ & $L_0$ & $L_1$ & $L_2$ & $L_2$ & $L_2$ & $L_2$\\
$L_3$ & $L_0$ & $L_1$ & $L_2$ & $L_3$ & $L_3$ & $L_3$\\
$L_4$ & $L_0$ & $L_1$ & $L_2$ & $L_3$ & $L_4$ & $L_4$\\
$L_5$ & $L_0$ & $L_1$ & $L_2$ & $L_3$ & $L_4$ & $L_5$\\
\hline
\end{tabular}
\caption{Multiplication table}
\label{CWTable4}
\end{table}

Let's consider now a simple two-source case with a 2D frame $\Theta=\{\theta_1,\theta_2\}$, Shafer's model for $\Theta$, and qba's expressed as follows:
$$qm_1(\theta_1)=L_1, \quad qm_1(\theta_2)=L_3, \quad qm_1(\theta_1\cup\theta_2)=L_1$$
$$qm_2(\theta_1)=L_2, \quad qm_2(\theta_2)=L_1, \quad qm_2(\theta_1\cup\theta_2)=L_2$$

\begin{itemize}
\item {\bf{Fusion with (qCR)}}: According to qCR combination rule \eqref{qCR}, one gets the result in Table \ref{CWTable5}, since
\begin{align*}
qm_{qCR}(\theta_1) &=qm_1(\theta_1)qm_2(\theta_1) + qm_1(\theta_1)qm_2(\theta_1\cup\theta_2)  + qm_2(\theta_1)qm_1(\theta_1\cup\theta_2)\\
&= (L_1\times L_2)+(L_1\times L_2)+(L_2\times L_1)\\
&= L_1+L_1+L_1 = L_{1+1+1}=L_3
\end{align*}
\begin{align*}
qm_{qCR}(\theta_2) &=qm_1(\theta_2)qm_2(\theta_2) + qm_1(\theta_2)qm_2(\theta_1\cup\theta_2)  + qm_2(\theta_2)qm_1(\theta_1\cup\theta_2)\\
&= (L_3\times L_1)+(L_3\times L_2)+(L_1\times L_1)\\
&= L_1+L_2+L_1 = L_{1+2+1}=L_4
\end{align*}
\begin{align*}
qm_{qCR}(\theta_1\cup\theta_2) & = qm_1(\theta_1\cup\theta_2)qm_2(\theta_1\cup\theta_2)= L_1\times L_2 = L_1
\end{align*}
\begin{align*}
qm_{qCR}(\emptyset) & \triangleq K_{12} =  qm_1(\theta_1)qm_2(\theta_2)+ qm_1(\theta_2)qm_2(\theta_1)\\
&= (L_1\times L_1) +  (L_2\times L_3) = L_1+L_2= L_3
\end{align*}
\end{itemize}
In summary, one gets
\begin{table}[!h]
\centering
\begin{tabular}{|l|ccccc|}
\hline
 & $\theta_1$ & $\theta_2$ & $\theta_1\cup\theta_2$ & $\emptyset$ & $\theta_1\cap\theta_2$\\
\hline
$qm_1(.)$ & $L_1$ & $L_3$ & $L_1$ &  &  \\
$qm_2(.)$ & $L_2$ & $L_1$ & $L_2$ &  &  \\
\hline
$qm_{qCR}(.)$ & $L_3$ & $L_{4}$ & $L_1$ &  $L_3$ &  $L_0$\\
\hline
\end{tabular}
\caption{Fusion with qCR}
\label{CWTable5}
\end{table}

\begin{itemize}
\item {\bf{Fusion with (qDSmC)}}: If we accepts the free-DSm model instead Shafer's model, according to qDSmC combination rule \eqref{qDSmC}, one gets the result in Table \ref{TableDSmC},
\end{itemize}

\begin{table}[h]
\centering
\begin{tabular}{|l|ccccc|}
\hline
 & $\theta_1$ & $\theta_2$ & $\theta_1\cup\theta_2$ & $\emptyset$ & $\theta_1\cap\theta_2$\\
\hline
$qm_1(.)$ & $L_1$ & $L_3$ & $L_1$ &  &  \\
$qm_2(.)$ & $L_2$ & $L_1$ & $L_2$ &  &  \\
\hline
$qm_{qDSmC}(.)$ & $L_3$ & $L_4$ & $L_1$ &  $L_0$ &  $L_3$\\
\hline
\end{tabular}
\caption{Fusion with qDSmC}
\label{TableDSmC}
\end{table}

\begin{itemize}
\item {\bf{Fusion with (qDSmH)}}: Working with Shafer's model for $\Theta$, according to qDSmH combination rule \eqref{qDSmH}, one gets the result in Table \ref{TableDSmH}.
\end{itemize}

\begin{table}[h]
\centering
\begin{tabular}{|l|ccccc|}
\hline
 & $\theta_1$ & $\theta_2$ & $\theta_1\cup\theta_2$ & $\emptyset$ & $\theta_1\cap\theta_2$\\
\hline
$qm_1(.)$ & $L_1$ & $L_3$ & $L_1$ &  &  \\
$qm_2(.)$ & $L_2$ & $L_1$ & $L_2$ &  &  \\
\hline
$qm_{qDSmH}(.)$ & $L_3$ & $L_4$ & $L_4$ &  $L_0$ &  $L_0$\\
\hline
\end{tabular}
\caption{Fusion with qDSmC}
\label{TableDSmH}
\end{table}

since $qm_{qDSmH}(\theta_1\cup\theta_2)=L_1+L_3=L_4$.

\begin{itemize}
\item {\bf{Fusion with (qPCR5)}}:
Following PCR5 method, we propose to transfer the qualitative partial masses
\begin{enumerate}
\item[a)] $qm_1(\theta_1)qm_2(\theta_2)=L_1\times L_1=L_1$ to $\theta_1$ and $\theta_2$ in equal parts (i.e. proportionally to $L_1$ and $L_1$ respectively, but $L_1=L_1$); hence $\frac{1}{2}L_1$ should go to each of them.
\item[b)] $qm_2(\theta_1)qm_1(\theta_2)=L_2\times L_3=L_2$ to $\theta_1$ and $\theta_2$ proportionally to $L_2$ and $L_3$ respectively; but since we are not able to do an exact proportionalization of labels, we approximate through transferring  $\frac{1}{3}L_2$ to $\theta_1$ and $\frac{2}{3}L_2$ to $\theta_2$.
\end{enumerate}

The transfer $(1/3) L_2$ to $\theta_1$ and $(2/3) L_2$ to $\theta_2$ is not arbitrary, but it is an approximation since the
transfer was done proportionally to $L_2$ and $L_3$, and $L_2$ is smaller than $L_3$; we mention that it is not possible to do an exact transferring. Nobody in the literature has done so far normalization of labels, and we tried to do a quasi-normalization\index{quasi-normalization} [i.e. an approximation].\\

Summing a) and b) we get: $\frac{1}{2}L_1 + \frac{1}{3}L_2\approx L_1$, which represents the partial conflicting qualitative mass\index{qualitative mass} transferred to $\theta_1$, and $\frac{1}{2}L_1 +  \frac{2}{3}L_2\approx L_2$, which represents the partial conflicting qualitative mass\index{qualitative mass} transferred to $\theta_2$. Here we have mixed qualitative and quantitative information.

Hence we will finally get:
\begin{table}[h]
\centering
\begin{tabular}{|l|ccccc|}
\hline
 & $\theta_1$ & $\theta_2$ & $\theta_1\cup\theta_2$ & $\emptyset$ & $\theta_1\cap\theta_2$\\
\hline
$qm_1(.)$ & $L_1$ & $L_3$ & $L_1$ &  &  \\
$qm_2(.)$ & $L_2$ & $L_1$ & $L_2$ &  &  \\
\hline
$qm_{qPCR5}(.)$ & $L_4$ & $L_5$ & $L_1$ &  $L_0$ &  $L_0$\\
\hline
\end{tabular}
\caption{Fusion with qPCR5}
\label{TableqPCR5}
\end{table}

\noindent
For the reason that we can not do a normalization (neither previous authors on qualitative fusion rules\index{qualitative fusion rules} did), we propose for the first time the possibility of {\it{quasi-normalization\index{quasi-normalization}}} (which is
an approximation of the normalization), i.e. instead of dividing each qualitative mass\index{qualitative mass} by a coefficient of normalization, we {\it{subtract}} from each qualitative mass\index{qualitative mass} a qualitative coefficient (label) of quasi-normalization\index{quasi-normalization} in order to adjust the sum of masses.\\

Subtraction on $L$ is defined in a similar way to the addition:
\begin{equation}
L_i - L_j=
\begin{cases}
L_{i-j}, &\quad\text{if} \ i\geq j;\\
L_0, &\quad\text{if} \ i< j;
\end{cases}
\label{qsub}
\end{equation}
\noindent
$L$ is closed under subtraction as well.\\

The subtraction can be used for quasi-normalization\index{quasi-normalization} only, i. e. moving the final label result 1-2 steps/labels up or down.  It is not used together with addition or multiplication.\\

The increment in the sum of fusioned qualitative masses\index{qualitative mass} is due to the fact that multiplication on $L$ is approximated by a larger number, because multiplying any two numbers $a$, $b$ in the interval $[0,1]$, the product is less than each of them, or we have approximated the product $a\times b = \min\{a,b\}$.

\noindent
Using the quasi-normalization\index{quasi-normalization} (subtracting $L_1$), one gets with qDSmH and qPCR5, the following {\it{quasi-normalized}} masses (we use $\star$ symbol to specify the quasi-normalization\index{quasi-normalization}):

\begin{table}[h]
\centering
\begin{tabular}{|l|ccccc|}
\hline
 & $\theta_1$ & $\theta_2$ & $\theta_1\cup\theta_2$ & $\emptyset$ & $\theta_1\cap\theta_2$\\
\hline
$qm_1(.)$ & $L_1$ & $L_3$ & $L_1$ &  &  \\
$qm_2(.)$ & $L_2$ & $L_1$ & $L_2$ &  &  \\
\hline
$qm_{qDSmH}^\star (.)$ & $L_2$ & $L_3$ & $L_3$ &  $L_0$ &  $L_0$\\
$qm_{qPCR5}^\star (.)$ & $L_3$ & $L_4$ & $L_0$ &  $L_0$ &  $L_0$\\
\hline
\end{tabular}
\caption{Fusion with quasi-normalization\index{quasi-normalization}}
\label{Tableqn}
\end{table}
\end{itemize}

\section{Conclusion}

A general presentation of foundation of DSmT has been proposed in this introduction which proposes new quantitative rules of combination for uncertain, imprecise and highly conflicting sources of information. Several applications of DSmT have been proposed recently in the literature and show the efficiency of this new approach over classical rules, mainly those based on the Demspter's rule in the DST framework. Recent PCR rules of combination (typically PCR no 5) have also been developed which offer a more precise transfer of partial conflicts than classical rules. DSmT rules have been also extented for the fusion of qualitative beliefs expressed in terms of linguistic labels for dealing directly with natural language and human reports. Matlab source code for the implementation of DSm rules and also new belief conditioning rules (not presented herein) have been recently developed and can be found in the forthcoming second DSmT book \cite{Book_2006}.


\begin{thebibliography}{99}

\bibitem{Bolanos_1993}
Bolanos J., De Campos L.M., Moral S., \emph{Propagation of linguistic labels in causal networks},
Proc. of 2nd IEEE Int. Conf. on Fuzzy Systems, Vol. 2, pp. 863--870, 28 March--1 April 1993.

\bibitem{Comtet_1974}
Comtet L., \emph{Sperner Systems},  sec.7.2 in Advanced Combinatorics: The Art of Finite and Infinite Expansions, D. Reidel Publ. Co., pp.Ê271-273, 1974. 

\bibitem{Dedekind_1897}
Dedekind R. \emph{†ber Zerlegungen von Zahlen durch ihre gršssten gemeinsammen Teiler}, In Gesammelte Werke, Bd. 1. pp.Ê103-148, 1897. 

\bibitem{Denoeux_1997}
Den\oe ux T, \emph{Reasoning with imprecise belief structures},
Technical Report Heudiasys 97/44, available at {\scriptsize{\verb+http://www.hds.utc.fr/~tdenoeux/+}}.

\bibitem{Denoeux_1999}
Den\oe ux T, \emph{Reasoning with imprecise belief structures},
International Journal of Approximate Reasoning, 20, pp. 79-111, 1999.

\bibitem{Dezert_2006a}
Dezert J., Smarandache F., \emph{DSmT: A New Paradigm Shift for Information Fusion}, in Proceedings of Cogis ' 06 Conference, Paris, March 2006.

\bibitem{Dezert_2006}
Dezert J., Tchamova A., Smarandache F., Konstantinova P., \emph{Target Type Tracking with PCR5 and Dempster's rules: A Comparative Analysis}, in Proceedings of Fusion 2006 International conference on Information Fusion, Fusion 2006, Firenze, Italy, July 10-13, 2006.

\bibitem{Dubois_1986c}
Dubois D., Prade H., \emph{On the unicity of Dempster rule of combination}, International Journal of Intelligent Systems, Vol. 1, pp 133-142, 1986.

\bibitem{Dubois_1988}
Dubois D., Prade H., \emph{Representation and combination of uncertainty with belief functions and possibility measures}, Computational Intelligence, 4, pp. 244-264, 1988.

\bibitem{Inagaki_1991}
Inagaki T., \emph{Interdependence between safety-control policy and multiple-sensor schemes via Dempster-Shafer theory}, IEEE Trans. on reliability, Vol. 40, no. 2, pp. 182-188, 1991.

\bibitem{Lamata_1994}
Lamata M., Moral S., \emph{Calculus with linguistic probabilities and beliefs},
In R. R. Yager, M. Fedrizzi, and J. Kacprzyk, editors, Advances in Dempster-Shafer Theory of Evidence, pp. 133-152, Wiley.

\bibitem{Lefevre_2002}
Lefevre E., Colot O., Vannoorenberghe P. \emph{Belief functions combination and conflict management}, Information Fusion Journal, Elsevier Publisher, Vol. 3, No. 2, pp. 149-162, 2002.

\bibitem{Lefevre_2003}
Lefevre E., Colot O., Vannoorenberghe P., \emph{Reply to the Comments of R. Haenni on the paper ''Belief functions combination and conflict management"}, 
Information Fusion Journal, Elsevier Publisher, Vol. 4, pp. 63-65, 2003.

\bibitem{Murphy_2000}
Murphy C.K.,\emph{Combining belief functions when evidence conflicts}, 
Decision Support Systems, Elsevier Publisher, Vol. 29, pp. 1-9, 2000.

\bibitem{Pearl_1988}
Pearl J., \emph{Probabilistic reasoning in Intelligent Systems: Networks of Plausible Inference}, Morgan Kaufmann Publishers, San Mateo, CA, 1988.

\bibitem{Robinson_1966}
Robinson A., \emph{Non-Standard  Analysis}, North-Holland Publ. Co., 1966.

\bibitem{Sentz_2002}
Sentz K., Ferson S., \emph{Combination of evidence in Dempster-Shafer Theory}, SANDIA Tech. Report, SAND2002-0835, 96 pages, April 2002.
\bibitem{Shafer_1976}
Shafer G., \emph{A Mathematical Theory of Evidence}, Princeton Univ. Press, Princeton, NJ, 1976.

\bibitem{Sloane_2003}
Sloane N.J.A., \emph{The On-line Encyclopedia of Integer Sequences 2003}, 
(Sequence No. A014466),
{\scriptsize{\verb+http://www.research.att.com/~njas/sequences/+}}.

\bibitem{Smarandache_2000}
Smarandache F., \emph{A Unifying Field in Logics: Neutrosophic Logic. Neutrosophy, Neutrosophic Set, Probability, and Statistics}, (2nd Ed.), Amer. Research Press, Rehoboth, 2000.

\bibitem{Smarandache_2002a}
Smarandache F., \emph{A Unifying Field in Logics: Neutrosophic Logic}, Multiple-valued logic, An international journal, Vol. 8, No. 3, pp. 385-438, 2002.

\bibitem{Smarandache_2002b}
Smarandache F., \emph{Neutrosophy: A new branch of philosophy}, Multiple-valued logic, An international journal, Vol. 8, No. 3, pp. 297-384, 2002.

\bibitem{Smarandache_2002}
Smarandache F. (Editor), \emph{Proceedings of the First International Conference on 
Neutrosophics}, Univ. of New Mexico, Gallup Campus, NM, USA, 1-3 Dec. 2001, Xiquan, Phoenix, 2002.

\bibitem{DSmTBook_2004a}
Smarandache F., Dezert J. (Editors), \emph{Applications and Advances of DSmT for Information Fusion}, Am. Res. Press, Rehoboth, 2004, 
http://www.gallup.unm.edu/{\verb+~+}smarandache/DSmT-book1.pdf. 

\bibitem{Smarandache_2005c}
Smarandache F., Dezert J., \emph{Information Fusion Based on New Proportional Conflict Redistribution Rules}, Proceedings of Fusion 2005 Conf., Philadelphia, July 26-29, 2005.

\bibitem{Book_2006}
Smarandache F., Dezert J. (Editors), \emph{Applications and Advances of DSmT for Information Fusion}, Vol. 2, American Research Press, Rehoboth, August 2006. 

\bibitem{Smets_1986}
Smets Ph., \emph{Combining non distinct evidence}, 
Proc. North American Fuzzy Information Processing (NAFIP 1986), New Orleans, LA, 1986.

\bibitem{Smets_1988}
Smets Ph.,Mamdani E.H., Dubois D., Prade H. (Editors), \emph{Non-Standard Logics for Automated Reasoning}, Academic Press, 1988.

\bibitem{Smets_1994}
Smets Ph., Kennes R., \emph{The transferable belief model}, 
Artif. Intel., 66(2), pp. 191-234, 1994.

\bibitem{Smets_2000}
Smets Ph., \emph{Data Fusion in the Transferable Belief Model}, 
Proceedings of the 3rd International Conference on Information Fusion, Fusion 2000, Paris, July 10-13, 2000, pp PS21-PS33.

\bibitem{Tombak_2001}
Tombak M., Isotamm A., Tamme T., \emph{On logical method for counting Dedekind numbers}, Lect.  Notes on Comp.Sci., 2138, p. 424-427, Springer-Verlag, 2001.
{\scriptsize{\verb+www.cs.ut.ee/people/m_tombak/publ.html+}}.

\bibitem{Voorbraak_1991}
Voorbraak F., \emph{On the justification of Dempster's rule of combination}, Artificial Intelligence, 48, pp. 171-197, 1991. (see \verb+http://turing.wins.uva.nl/~fransv/#pub+).

\bibitem{Yager_1983}
Yager R. R., \emph{Hedging in the combination of evidence}, Journal of Information and Optimization Science, Vol. 4, No. 1, pp. 73-81, 1983.

\bibitem{Yager_1985}
Yager R. R., \emph{On the relationships of methods of aggregation of evidence in expert systems}, Cybernetics and Systems, Vol. 16, pp. 1-21, 1985.

\bibitem{Yager_1987}
Yager R.R., \emph{On the Dempster-Shafer framework and new combination rules}, Information Sciences, Vol. 41, pp. 93--138, 1987.

\bibitem{Zadeh_1965}
Zadeh, L., \emph{Fuzzy sets}, Inform and Control 8, pp. 338-353, 1965.

\bibitem{Zadeh_1975}
Zadeh, L., \emph{Fuzzy Logic and Approximate Reasoning}, Synthese, 30, 407-428, 1975.

\bibitem{Zadeh_1979}
Zadeh L., \emph{On the validity of Dempster's rule of combination}, Memo M 79/24, Univ. of California, Berkeley, 1979.

\bibitem{Zadeh_1984}
Zadeh L., \emph{Review of Mathematical theory of evidence, by Glenn Shafer}, AI Magazine, Vol. 5, No. 3, pp. 81-83, 1984.

\bibitem{Zadeh_1985}
Zadeh L., \emph{A simple view of the Dempster-Shafer theory of evidence and its implications for the rule of combination}, Berkeley Cognitive Science Report No. 33, University of California, Berkeley, CA, 1985.

\bibitem{Zadeh_1986}
Zadeh L., \emph{A simple view of the Dempster-Shafer theory of evidence and its implication for the rule of combination}, AI Magazine 7, No.2, pp. 85-90, 1986.

\bibitem{Zhang_1994}
Zhang L., \emph{Representation, independence, and combination of evidence in the Dempster-Shafer Theory}, Advances in the Dempster-Shafer Theory of Evidence, R.R. Yager, J. Kacprzyk and M. Fedrizzi, Eds., John Wiley and Sons, Inc., New York, pp. 51-69, 1994.
\end{thebibliography}
\end{document}